\documentclass {article} 
\usepackage{arxiv}
\usepackage{booktabs} 
\usepackage{algorithm}%
\usepackage{algpseudocode}%
\usepackage{hyperref}
\usepackage{amsmath}
\usepackage{xcolor}
\usepackage{xspace}
\usepackage{soul}
\newcommand{\ignore}[1]{}
\usepackage[textsize=tiny]{todonotes}
\usepackage{graphicx}
\usepackage{boldline}
\usepackage[export]{adjustbox}
\usepackage[caption=false]{subfig}
\usepackage[font=small]{caption}
\usepackage{adjustbox}
\usepackage{enumerate}
\usepackage{amsfonts}
\usepackage{colortbl}
\usepackage{float}
\usepackage{rotating}
\usepackage{setspace}
\usepackage{textcomp}
\usepackage{gensymb}	
\usepackage[official]{eurosym} 
\usepackage[flushleft]{threeparttable}
\usepackage{enumitem}
\usepackage{array}
\title{Optimisation of Large Wave Farms using a Multi-strategy Evolutionary Framework}

\author{
 Mehdi Neshat \\
  Optimization and Logistics Group\\
  School of Computer Science\\
  The University of Adelaide\\
   Australia \\
  \texttt{mehdi.neshat@adelaide.edu.au} \\
   \And
 Bradley Alexander \\
  Optimization and Logistics Group\\
  School of Computer Science\\
  The University of Adelaide\\
   Australia \\
  \texttt{bradley.alexander@adelaide.edu.au} \\
  \And
   Nataliia Y. Sergiienko\\
School of Mechanical Engineering\\
	The University of Adelaide\\
	 Australia\\
	 \texttt{nataliia.sergiienko@adelaide.edu.au} \\
   \And
 Markus Wagner \\
  Optimization and Logistics Group\\
  School of Computer Science\\
  The University of Adelaide\\
   Australia \\
  \texttt{markus.wagner@adelaide.edu.au} \\
  }

\begin{document}

\maketitle
\doublespacing
\begin{abstract}

Wave energy is a fast-developing and promising renewable energy resource.
The primary goal of this research is to maximise the total harnessed power of a large wave farm consisting of fully-submerged three-tether wave energy converters (WECs).  
Energy maximisation for large farms is a challenging search problem due to the costly calculations of the hydrodynamic interactions between WECs in a large wave farm and the high dimensionality of the search space.
To address this problem, we propose a new hybrid multi-strategy evolutionary framework combining smart initialisation, binary population-based evolutionary algorithm, discrete local search and continuous global optimisation. 
For assessing the performance of the proposed hybrid method, we compare it with a wide variety of state-of-the-art optimisation approaches, including six continuous evolutionary algorithms, four discrete search techniques and three hybrid optimisation methods. 
The results show that the proposed method performs considerably better in terms of convergence speed and farm output. 
\end{abstract}


\sloppy

\section{Introduction}
The use of renewable energy sources continues to exhibit very fast growth of deployment, and it has resulted in savings of more than two gigatonnes of carbon dioxide in 2018 alone~\cite{ajadi2019global}.  
One of the most promising renewable sources is ocean wave energy, which has a high energy density per unit area of ocean, high level of predictability, and potentially high capacity factors~\cite{drew2009review,vicinanza2013wave}. 
However, compared to wind and solar energy, wave energy is still a nascent field, and research is still very active converter design~\cite{sergiienko2020design}, wave-farm layout, and power-take-off parameters~\cite{cuadra2016computational, de2014factors}.   

While there has been significant research on the placement of wave energy converters (WECs) in farms~\cite{child2010optimal,ruiz2017layout,wu2016fast,neshat2018detailed,neshat2019new}, to date, only Wu et al.~\cite{wu2016fast} has considered the design of larger layouts of over 20 converters, using a much-simplified wave energy model. 
 
The research described in this paper extends previous work by using a much more detailed energy model to place buoys in large farms of up to 100 WECs. Due to the much higher number of interactions modelled in such farms this work requires the development of novel, specialised, and highly-efficient search heuristics. Using an improved energy model, we demonstrate the performance of these new algorithms in two contrasting real wave scenarios (Sydney and Perth) and compare their performance to a suite of extant optimisation algorithms.  

This paper is organised as follows. In the next section, we survey related work. Section~\ref{sec:model} describes our WEC model. Section~\ref{sec:opt} formulates the optimisation problem. The proposed optimisation methods are described in Section~\ref{sec:method}.
The results of the optimisation experiments, including simple landscape analysis, are described in~Section~\ref{sec:experiments}. Section~\ref{sec:conclusions} concludes this paper and canvases future work. 

\section{Related Work}

Placement of WECs in larger farms is a challenging optimisation problem. Hydrodynamic interactions between WECs are complex, which makes evaluation of each potential layout time-consuming~\cite{neshat2018detailed}, ranging from minutes to hours for large farms. Second, due to complex inter-WEC interactions, the search space for this problem is multi-modal -- thus requiring global search to be assured of good results. Finally, the high number of decision variables in large farms increases the search space to traverse. 

There has been substantial past research into the problem of WEC placement. One of the first studies to optimise WEC layout compared a customised genetic algorithm (GA) with an iterative Parabolic Intersection (PI) method~\cite{child2010optimal} for a small wave farm (five buoys). The GA outperformed PI, but required more evaluations to do so.  
A more recent position optimisation study~\cite{ruiz2017layout} compared  three search metaheuristics:  a custom GA, CMA-ES~\cite{hansen2006cma}, and glow-worm optimisation~\cite{krishnanand2009glowworm}), using a simple wave model. The study observed that CMA-ES converges the fastest, while the other models produced slightly better results. 
Wu et al. \cite{wu2016fast} considered optimising a large wave farm (25--100 WECs) as an array of fully submerged three-tether buoys using 1+1EA and 2+2CMA-ES. That research found that the 1+1EA with a simple mutation operator performed better than CMA-ES. A limitation of that work was that it was limited to a highly simplified single-wave-direction wave scenario. 

In a move toward problem-specific algorithms, Neshat et al.~\cite{neshat2018detailed} proposed a hybrid optimisation method (LS-NM) combined with a neighbourhood search and Nelder-Mead search. Their study found that LS-NM performed better than generic and custom EAs. However, the wave model applied by that study, though quite detailed,  still used an artificial wave scenario and small farm sizes (4 and 16 WECs). 
More recently, more problem-specific search techniques~\cite{neshat2019new,neshat2019hybrid} were, respectively,  proposed for optimising WECs positions by utilising a surrogate power model (that is learned on the fly); and hybrid symmetric local search by defining a search sector to speed up the optimisation process. These approaches were also applied to real wave scenarios.
For handling this real expensive optimisation problem, a neuro-surrogate optimisation approach was recommended~\cite{neshat2019adaptive} that is composed of a surrogate Recurrent Neural Network (RNN) model and a symmetric local search.  This surrogate model is joined with a metaheuristic (Gray Wolf optimiser) for tuning the model’s hyper-parameters. 
However, these search strategies performance were not evaluated on a large farm. 

This article differs from previous work by optimising large layouts using an improved high-fidelity hydrodynamic model to optimise layouts in real wave scenarios. 
We develop a new hybrid multi-strategy evolutionary algorithm for optimising the positions of buoys in the wave farm to maximise the average total farm power output. For evaluating the new algorithm, we compare its performance to: (1) six continuous off-the-shelf evolutionary methods, (2) four discrete heuristic approaches (3 new), population and individual-based, and (3) three new hybrid EAs (continuous+discrete). We use these methods to optimise wave farms of sizes 49 and 100. We use fine-grained models of contrasting real wave climates, Perth and Sydney, which are located off the southern coast of Australia. The optimisation results demonstrate that the new hybrid multi-strategy search approach produces the best results.

\section{The wave energy converter model} \label{sec:model}

This section describes the energy model for WEC layouts used in this study. The WEC design simulated here is a three-tether spherical buoy based on the highly effective CETO 6 system developed by Carnegie Clean Energy~\cite{mann2007ceto}.  

\subsection{Equation of motion}

We model a fully submerged spherical buoy of 5 m radius that is tethered to three power take-off units installed on a seabed. A detailed description of this WEC and its physical parameters can be found in~\cite{neshat2018detailed}.

The motion of each buoy in the farm depends on the forces due to the fluid-structure interaction and the force exerted on the buoy from the PTO system. The generalised equation that describes the motion of all buoys can be written in the frequency domain as:
\begin{equation}
    (\mathbf{M}+\mathbf{A})\ddot{\mathbf{X}} + (\mathbf{B}+\mathbf{D}_{pto})\dot{\mathbf{X}} + \mathbf{K}_{pto}{\mathbf{X}} = \mathbf{F}_{exc}, \label{eq:motion}
\end{equation}
where $\mathbf{X}\in \mathbb{R}^{3N\times 1}$ is a vector of surge, sway and heave displacements of each buoy, $\mathbf{M} = m \mathbb{I}_{3N}$ is a diagonal mass matrix of the wave farm, $\mathbf{A}$ and $\mathbf{B}\in \mathbb{R}^{3N\times 3N}$ are the matrices of hydrodynamic added mass and damping coefficients respectively, $\mathbf{K}_{pto}$ and $\mathbf{D}_{pto}\in \mathbb{R}^{3N\times 3N}$ are the block diagonal matrices of PTO stiffness and damping coefficients respectively, and $\mathbf{F}_{exc} \in \mathbb{R}^{3N\times 1}$ is a vector of excitation forces.

\subsection{Performance assessment}
After solving the equation of motion \eqref{eq:motion}, we can calculate the power absorbed by the farm in a regular wave of frequency $\omega$ that propagates from direction $\beta$:
\begin{equation}
    p(\omega, \beta) = \frac{1}{2}\dot{\mathbf{X}}^{*}\mathbf{D}_{pto}\dot{\mathbf{X}} \label{eq:power_freq}
\end{equation}
where $()^*$ denotes the conjugate transpose of a matrix.

Eq.~\eqref{eq:power_freq} allows us to estimate the power production of a farm assuming that the ocean wave has only one frequency component (like a sinusoidal wave) and propagates only from one direction. In reality, ocean waves travel from different directions and contain multiple frequencies. This behaviour of the wave is usually described by the directional wave spectrum $S(\omega, \beta)$, and power generated by the wave farm in the irregular wave, or sea state $(H_s, T_p)$, can be approximated by:
\begin{equation}
    P(H_s, T_p) = \int_0^{2\pi} \int_0^{\infty} S(\omega, \beta) p(\omega, \beta)\,d\omega \,d\beta. \label{eq:power_ss}
\end{equation}

A potential deployment site (e.g. Perth or Sydney) can be characterised by the wave climate where each sea state has the probability of occurrence $O(H_s, T_p)$. Therefore, using values from Eq.~\eqref{eq:power_ss} and having historical wave climate statistics, it is possible to calculate the annual average power generated by the wave farm at a given location:
\begin{equation}
    P_{\Sigma} = \sum  P(H_s, T_p) O(H_s, T_p).
\end{equation}

The  Perth and Sydney sites are qualitatively very different: Perth has a small sector from which the prevailing waves arrive, while Sydney's wave directions vary much more. For Perth, this can result in very pronounced constructive and destructive interference, while the same are ``smeared'' out for Sydney, thus resulting in two very different optimisation scenarios.

Another metric that is widely used to demonstrate the quality of the buoy placement in a farm is called the $q$-factor. It can be calculated as a ratio of the power generated by the entire farm $P_{\Sigma}$ to the sum of power outputs from all WECs if they operate in isolation (not in a farm) $P^i_{\Sigma}$:
\begin{equation}
    q = \frac{P_{\Sigma}}{\sum_i^N P^i_{\Sigma}}
\end{equation}
Values of $q>1$ indicate that this particular farm benefits from the constructive interaction between WECs, and more energy can be generated if these WECs operate together.  

The MATLAB implementation of this model can be downloaded at \cite{sergiienko2020simulator}.


\section{Optimisation problem formulation}\label{sec:opt}

Based on our WEC model, the problem of positioning $N$ converters on a restricted area of a wave farm $(l\times w)$ in order to maximise the average
annual power production $P_{\Sigma}$ is:
\[
  P_{\Sigma}^* = \mbox{\em argmax}_{\mathbf{x,y}} P_{\Sigma}(\mathbf{x,y})
\]
\noindent where $P_{\Sigma}(\mathbf{x,y})$ is the average power obtained by placements of the buoys in a field at $x$-positions $\mathbf{x}=[x_1,\ldots,x_N]$ and corresponding $y$ positions $\mathbf{y}=[y_1,\ldots,y_N]$. In our experiments, the number of buoys is $N=49$ and $100$. 

\paragraph{Constraints}
All buoy positions $(x_i,y_i)$ are constrained to a square field of dimensions: $l\times w$ where $l=w=\sqrt{N * 20000}\,m$. This allocates $20000m^2$ of farm-area per-buoy. In addition, the intra-buoy distance must not be less than 50 meters for reasons of safety and maintenance access. For any layout $\mathbf{x,y}$ the sum-total of the inter-buoy distance violations, measured in metres, is:

\vspace{2mm}\hspace{5mm}$\mbox{\em{Sum}}_{\mbox{\em dist}}= \sum_{i=1}^{N-1}\sum_{j=i+1}^{N} 
(\mbox{\em{dist}}((x_i,y_i),(x_j,y_j))-50), $

\hspace{28mm}$\mbox{if } \mbox{\em{dist}}((x_i,y_i),(x_j,y_j))<50$ \mbox{else 0}

\vspace{2mm}\noindent where $\mbox{\em dist}((x_i,y_i),(x_j,y_j))$ is the Euclidean distance between each pair of buoys $i$ and $j$.

Violations of the inter-buoy distance constraint are handled by applying a steep penalty function: $(\mbox{\em{ Sum}}_{\mbox{\em{dist}}}+1)^{20}$ and then applying the Nelder-Mead simplex algorithm over this penalty function to repair the violations in the layout.
This approach avoids expensive re-evaluations of the full-wave model that would be required if the penalty function were combined with the full model whilst repairing distance violations. 
Meanwhile, we handle buoy placements outside of the farm area by moving them back to the farm boundary. 

\paragraph{Computational Resources}

In this paper, we aim to compare several heuristic search methods, for 49 and 100-buoy layouts, in two realistic wave models. Because the search methods apply the interaction model to differing numbers of buoys at a time, it is not feasible to compare methods fairly in terms of a fixed number of model evaluations. 

Instead, we use an allocated time budget for each run of three days on dedicated nodes of an HPC platform with 2.4GHz Intel 6148 processors and 128GB of RAM. 
The software environment running the function evaluations and the search algorithms is MATLAB R2019. On this platform, 12-fold parallelisation inside of Matlab yields up to 10-fold speedup. All algorithm variants are carefully implemented to make use of the parallelism available. 

\section{Optimisation Methods}\label{sec:method}
The algorithms that follow apply three broad strategies.
In the first strategy, we optimise in a continuous space using five off-the-shelf evolutionary algorithms. We also use the LS-NM \cite{neshat2018detailed} algorithm, which places and fine-tunes one buoy at a time. 
In the second strategy, we optimise the positions in a discretised grid where the spacing is based on the safety-distance. Here, we consider four different EAs.

Last, we propose a hybrid multi-strategy heuristic that is designed based on our observations that attempts to combine the strengths of the algorithms from the first two strategies.

\subsection{Continuous methods}
For the continuous optimisation strategy, we compare six meta-heuristic approaches to optimise all problem dimensions simultaneously: 
\begin{enumerate}
  \item covariance matrix adaptation evolutionary-strategy (CMA-ES) \cite{hansen2006cma,Hanson2009CMAES} which is an state-of-the-art and self-adaptive EA with the default $\lambda=12$, and initial $\sigma=0.25\times(U_b-L_b)$;
  \item (2+2)CMA-ES \cite{wu2016fast} with the default $\lambda=2$, and $\sigma=0.3\times(U_b-L_b)$;
  \item Differential Evolution (DE)~\cite{storn1997differential}, a well known global search heuristic using a binomial crossover and a mutation operator of $DE/rand/1/bin$,  The population size is adjusted by the $\lambda=12$ and other control parameters are $F=0.5$, $P_{cr}=0.8,0.9$ respectively for 49 and 100-buoy layouts;
  \item Improved  Differential Evolution \cite{fang2018optimization}, 
  with $\lambda=12$, and generating mutation vector in the form of $DE/best/1/bin$ with an adaptive mutation operator $F=F_0\times2^{e^{1-\frac{G_m}{G_{m+1}-G}}}$, where $F_0=0.5$ and $G_m$ is the maximum number of generations and $G$ is the current generation;
  \item a simple (1+1)EA as used in~\cite{wu2016fast} that mutates one buoy location in each iteration with a probability of $1/N$ using a normal distribution ($\sigma=0.1\times(U_b-L_b)$);
  \item Local Search + Nelder-Mead (LS-NM) \cite{neshat2018detailed}: which is a fast and effective WEC position optimisation method. Each buoy is placed and optimised one-at-a-time sequentially by sampling at a normally-distributed random offset ($\sigma=70m$) from the earlier placed buoy position. The sampled position proffering the highest power output, after NM search, is taken.
\end{enumerate}

\subsection{Discrete methods}
We test and compare four discrete optimisation methods. All methods place buoys at locations on a grid spaced at the safety distance of 50m. A-priori this discretisation offers advantages in terms of avoiding infeasible layouts and reduced overall search space. 
 
The discrete algorithms used here are:
\begin{enumerate}
\item binary Genetic Algorithm (bGA)~\cite{sharp2018wave} with $\lambda=12$, ~$e_p=10\%, ~C_r=80\%, ~M_r=10\%$, a binary mutation and double point crossover with respect to the number of buoys as a constraint, where $e_p,~C_r$ and $M_r$ are the elitism, crossover and mutation rate respectively. 
\item Improved binary Differential Evolution (bDE)~\cite{zorarpaci2016hybrid} with the same IDE settings and to construct the mutant vector, formula \ref{eq:difference_vector} and \ref{eq:mutant_vector} are used;
\begin{equation} \label{eq:difference_vector}
\mathit{Diff-Vector^j}=
\begin {cases}
0, & if(X_{r1}^j=X_{r2}^j)\\
X_{r1}^j, & \mathit{otherwise}
\end{cases}
\end{equation}
\begin{equation} \label{eq:mutant_vector}
\mathit{Mutant-Vector^j}=
\begin {cases}
1, & if(\mathit{Diff-Vector}^j=1)\\
X_{G_{best}}^j, & \mathit{otherwise}
\end{cases}
\end{equation}
 where $r_1$ and $r2$ are the index of two randomly chosen individuals, and $G_{best}$ mentions the best solution number in the current population. 
\item Enhanced binary Particle Swarm Optimisation (bPSO)~\cite{mirjalili2013s,Mirjalili2020bPSO} with $\lambda=12$, and other settings are $C_1=C_2=2$, $\omega_{ini}=2, \omega_{max}=0.9, \omega_{min}=0.4$, and $\omega$ is linearly decreased to $1.5$.
 the applied transfer function (V-shaped) is represented by Equation \ref{eq:transfer_function} and the position vector is updated by Equation~\ref{eq:position_updated}.
 \begin{equation} \label{eq:transfer_function}
T(v_i^k(t))=\left|\frac{2}{\pi}\arctan (\frac{\pi}{2}v_i^k(t))\right|
\end{equation}
\begin{equation} \label{eq:position_updated}
X_i^k(t+1)=
\begin {cases}
(X_i^k(t))^{-1} & if ~\mathit{rand} < T(v_i^k(t))\\
X_i^k(t) & \mathit{otherwise}
\end{cases}
\end{equation}
 where $v_i^k(t)$ indicates the $i^{th}$ particle velocity at iteration $t$ in the $k^{th}$ dimension. 
\item Discrete Local Search (DLS), which is an individual-based evolutionary algorithm similar to a $1+1EA$ with two kinds of mutation step sizes: one discrete interval based on a uniform random distribution that can be vertical, horizontal or diagonal with the same probability, and using a discrete normally-distributed random offset with $\sigma=3$ (DLS(II)). The mutation probability rate is $\frac{1}{N}$, where $N$ is the number of buoys. In DLS(II), we first generate an initial population, and then the best arrangement is chosen as a start individual. 
\end{enumerate}
\begin{figure}[t]
    \centering
 \includegraphics[height=7cm]{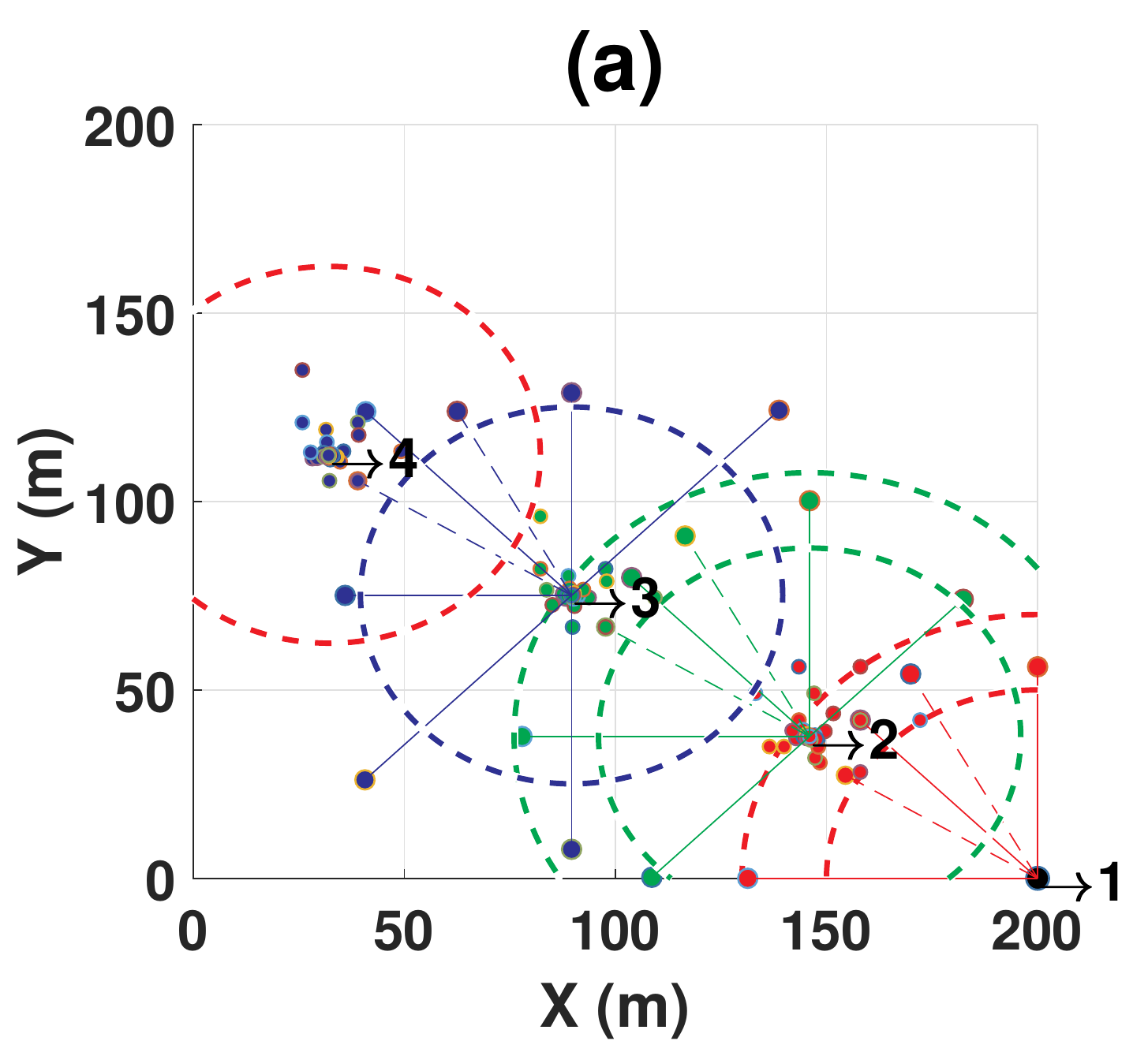}
  \includegraphics[height=7cm]{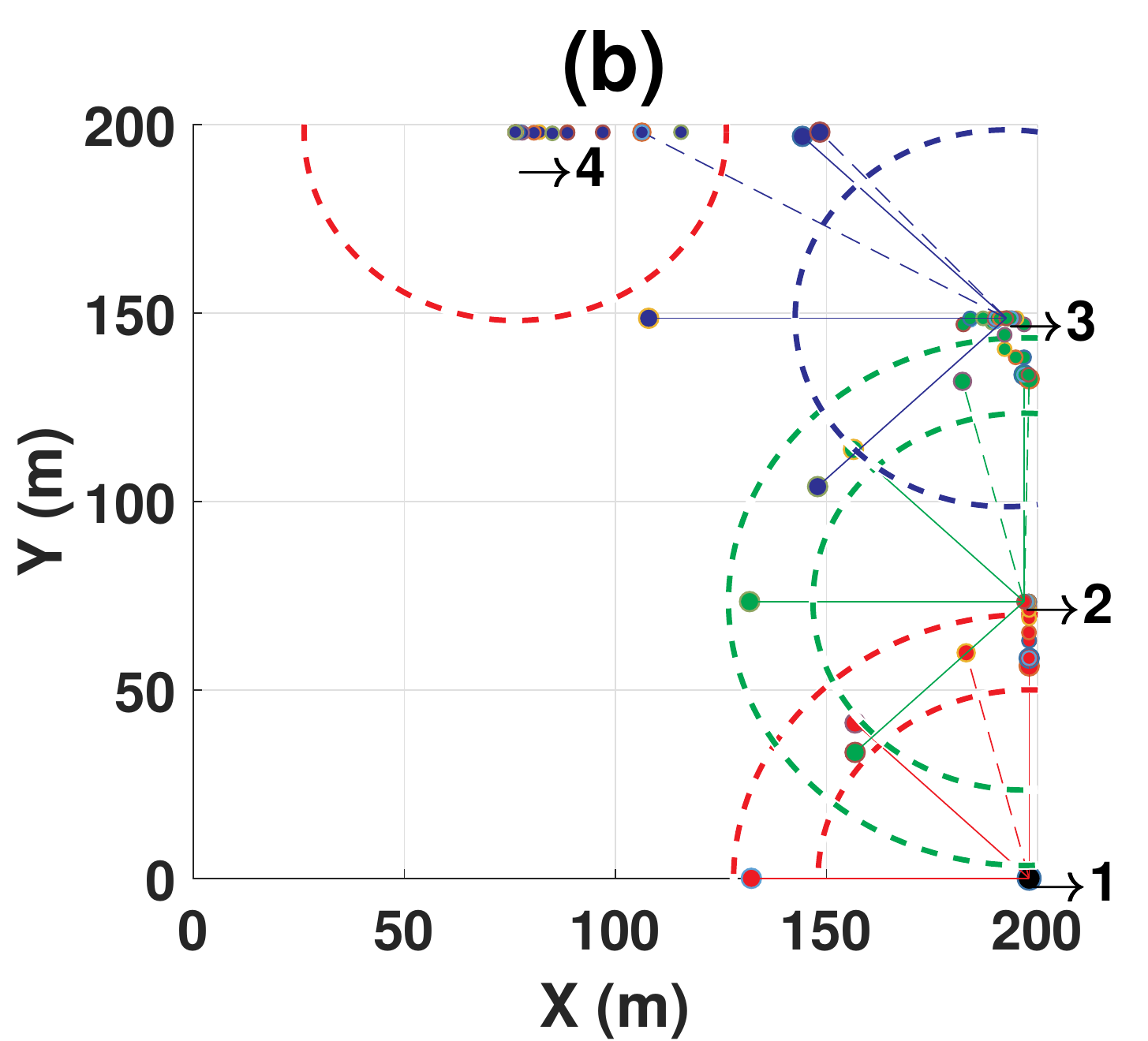}
     \caption{Symmetric Local Search + Nelder-Mead (SLS-NM)~\cite{neshat2019hybrid} for making the surrogate model. (a) 4-buoy layout in Perth, power=399474 (Watt), (b) Sydney wave model, power=405943 (Watt). The order of the placed and modified buoys position are numbered. }
    \label{fig:surrogate_model}
\end{figure}
\subsection{Hybrid methods}
\label{sec:Hybrid}
In the third strategy, we proposed a hybrid heuristic framework which  consists of five steps:

\begin{itemize}
\item \textbf{First step}: Applying Symmetric Local Search + Nelder-Mead (SLS-NM)~\cite{neshat2019hybrid}: which places one buoy at a time but offers a more systematic local search. The search starts by placing the first buoy in a pre-determined position of the field based on the dominant wave direction; a symmetric sampling around the neighbourhood of the current buoy is done. Next, Nelder-Mead is applied to optimise the placed buoy arrangement concerning the continuous variables. This process is repeated until the last buoy is placed. In our present case, we use SLS-NM to optimise a local 4-buoy sub-layout which will act as a surrogate layout model. Restricting the model to interactions between just 4 buoys makes these evaluations very fast and efficient. Figure~\ref{fig:surrogate_model} shows the detailed behaviour of this step. 
\item \textbf{Second step}: Discretising the search space (wave farm) based on the size of the surrogate sub-layout model as a smart initialisation method. Thus composing a large wave farm as a mosaic of the small surrogate sub-layouts that produce the most energy.

\item \textbf{Third step}: Generating the initial population with a sufficient number of well-arranged 4-buoy sub-layouts (smart initialisation) and then encoding to binary representation in preparation for running binary GAs on WEC positions.
\item \textbf{Fourth step}: Applying discrete optimisation methods on binary representations. We evaluate and compare the performance of three methods (bDE~\cite{zorarpaci2016hybrid}, bGA ~\cite{sharp2018wave} and bPSO~\cite{mirjalili2013s}).

\item \textbf{Fifth step}: if the improvement rate of the last populations of the applied optimisation method is low, the rotate procedure is run to perturb sub-layouts and avoid premature convergence. The rotate algorithm mutates a  4-buoy sub-layout by a random clockwise rotation degree with discrete 45\textdegree~ intervals.

The probability of the applied rotation on each sub-layout is $\frac{1}{N}$.

\end{itemize}

Using this configurable method, we compare three combinations:
\begin{enumerate}
\item SLS-NM + binary GA + Rotate (SLSNM-bGA)
\item SLS-NM + Improved binary DE + Rotate (SLSNM-bDE)
\item SLS-NM + Enhanced binary PSO + Rotate (SLSNM-bPSO)
\end{enumerate}

\subsection{Hybrid Multi-strategy Evolutionary algorithms }

The binary-encoded search space in the third hybrid search strategy is discrete. This means that, often, there is still scope to further tune layout locations. To implement this tuning we develop the third hybrid search strategy using a backtracking method for enhancing the buoys position. This backtracking idea is consists of 
\begin{enumerate}
\item A discrete local search (DLS) for providing a second chance for running a fast neighbourhood exploration of the buoys with a large step size (interval=50m) 

\item  and a continuous local search (CLS) that uses a 1+1EA for exploring near each buoy using small random normally distributed step size ($\sigma=20m$, linearly decreased).
\end{enumerate}
 Furthermore, the rotation procedure is embedded with the discrete metaheuristic algorithms as a mutation operator which is applied to perturb the best solution after each generation. According to the above descriptions, three Hybrid Multi-strategy Evolutionary algorithms are proposed including 
 \begin{enumerate}
\item SLS-NM + bGA-Rotate + DLS + CLS (MS-bGA)
\item SLS-NM+Improved bDE-Rotate + DLS + CLS (MS-bDE)
\item SLS-NM+Enhanced bPSO-Rotate + DLS + CLS (MS-bPSO)
\end{enumerate}
Algorithm \ref{alg:MS-bDE} describes MS-bDE in detail, where $N$, $N_s$, $N_b$ are the buoy numbers, the surrogate model's buoy number (4-buoy layout) and the number of binary decision variables respectively. And also both $Tr_1$ and $Tr_2$ are the stopping criteria of 24 (hours) and 48 (hours) respectively.  

\section{Experimental study}\label{sec:experiments}

This section shows detailed optimisation results comparing the 17 variations of search heuristics (six existing methods with and 11 new combinations) described in the previous section. In order to evaluate the performance of the proposed algorithms, we performed a comparative study using two distinct real wave scenarios (Perth and Sydney), and for two different large farm sizes with $N=49$ and $N=100$ buoys. For each optimisation method with the configurations above, we execute ten runs. For a set of runs, we tracked performance distributions, and the best layouts were gathered to compare each method. 

Table~\ref{table:allresults} shows summary statistics from the experimental runs. The best-obtained results are indicated in bold type. The minimum, maximum, average, median and standard deviation (STD) of the best-produced solutions (power output) for each experiment are reported. 
\begin{algorithm}[H]\small
\caption{$\mathit{MS-bDE}$}\label{alg:MS-bDE}
\begin{algorithmic}[1]
\Procedure{Hybrid Multi-strategy Evolutionary Algorithm}{}\\
 \textbf{Initialisation}
 \\$\mathit{N}=49, 100$, $\mathit{N_s}=4$, $\mathit{N_b}=N/N_s$,$\mathit{N_{Pop}}=12$, $\mathit{F_0}=0.5$, $\mathit{P_{cr}}=0.9, \mathit{iter}=1$
 \\$\mathit{size}=\sqrt{N*20000}$  \Comment{Farm size}
 \\$\vec{\mathit{\vec{S_s}}}=\{\langle x_1,y_1 \rangle,\ldots,\langle x_{N_s},y_{N_s}\rangle\}=\bot$\Comment{Continuous surrogate position}
 
 \\$\vec{\mathit{S}}=\{\langle x_1,y_1 \rangle,\ldots,\langle x_N,y_N \rangle\}=\bot$ \Comment{Discrete layout position}
 
 \\$\mathit{\chi_{dis}}=\{\langle \vec{S_1} \rangle,\langle \vec{S_2}\rangle\ldots,\langle \vec{S}_{N_{Pop}} \rangle\}$ \Comment{Discrete Population}
 
 \\ \textbf{Symmetric Local Search + Nelder-Mead (SLS-NM)}
\\ ($\mathit{energy_s,Array_s})=\mathbf{SLS-NM}([\vec{S_s}])$ \Comment{Optimise surrogate model}

\\{$\mathit{\chi^{iter}_{dis}}$=${\mathbf{IniFirstPop}}(\mathit{Array_s,\chi_{dis}})$}\Comment{Generate initial discrete population}

\\($\mathit{Energy,bestEnergy,bestArray})=\mathbf{Eval}({\mathit{\chi^{iter}_{dis}}})$\Comment{Evaluate population}
\\{$\mathit{\chi^{iter}_{b}}$=${\mathbf{ConDisBin}}(\mathit{\chi_{dis}^{iter}},N_b)$}\Comment{Encode discrete to binary population}

\\ \textbf{Discrete Differential Evolution (bDE)} 
\While{$\mathit{ImPorate\ge 0.1\%}~\&~ \mathit{\sum_{t=1}^{iter} runtime_t \le Tr_1}$}
 
 \For{ $i$ in $[1,..,N_{Pop}]$ } \Comment{ Mutation}
\State Generate two rand indexes $r_1,r_2\in(1,N_{Pop})$, $r_1\ne r_2 \ne i$ 
\State Compute mutant vector ($V_i^{iter}$) by Equations. \ref{eq:difference_vector} and \ref{eq:mutant_vector}
  
 \For{$j$ in $[1,..,\mathit{N_b}]$} \Comment{Crossover}
    \If{ {\em{$rand \le P_{cr}$ or $j==j_{rand}$}} }
       \State $U_{i,j}^{iter}=V_{i,j}^{iter}$
     \Else  
      \State $U_{i,j}^{iter}=\chi_{b_{i,j}}^{iter}$
  \EndIf
  \EndFor
   \If{ $f(U_i^{iter})$ $\ge$ $f(\chi_{b_i}^{iter})$ } \Comment{Selection(Maximisation)}
       \State $\chi_{b_{i}}^{iter+1}=U_{i}^{iter}$
     \Else  
      \State $\chi_{b_{i}}^{iter+1}=\chi_{b_{i}}^{iter}$
  \EndIf
  \EndFor 
  \State  ($\mathit{bestarray, bestIndex,bestEnergy, ImPorate}$)=$\mathbf{Max}(\chi_b^{iter+1}$)
   
  \\\textbf{\em{Rotation Operator}}
  
  \For{$k$ in $[1,..,\mathit{N_b}]$} 
  \If{$rand<\frac{1}{N_b}$}
  \State  ($\mathit{array_{R_k}}$)=$\mathbf{Rotate}(\mathit{bestarray,k}$)
  \EndIf
  \EndFor
  
  \State ($\mathit{Energy_R})=\mathbf{Eval}({\mathit{array_{R}}})$\Comment{Evaluate rotated layout}
  
  \State $\mathit{\chi_{b_{bestIndex}}^{iter+1}}= 
  \begin{cases}
  \mathit{array_{R}},& \text{\small if $\mathit{Energy_{R}>bestEnergy}$}  \\
  \mathit{bestarray},& \text{Otherwise}\\
    \end{cases}
  $\;
  \State $\mathit{iter=iter+1}$, and Update $\mathit{ImPorate}$
  \EndWhile 
  \State  ($\mathit{bestarray, bestIndex,bestEnergy, ImPorate}$)=$\mathbf{Max}(\chi_b^{iter}$)
  
\\ \textbf{\em{Discrete Local Search}}
  \While{$\mathit{ImPorate\ge 0.001\%}~\&~ \mathit{\sum_{t=1}^{iter} runtime_t\le Tr_2}$}
  \State ($\mathit{array_{dls},Energy_{dls}})=\mathbf{DLS}({\mathit{bestarray}})$
  
  \If{$\mathit{Energy_{dls}>bestEnergy}$  } 
       \State $\mathit{bestarray}$ = $\mathit{array_{dls}}$
       \State $\mathit{bestEnergy}$ = $\mathit{Energy_{dls}}$
       \State Update $\mathit{ImPorate}$
  \EndIf

  \EndWhile
\\ \textbf{\em{Continuous Local Search}} 
\While{$\mathit{\sum_{t=1}^{iter} runtime_t\le 72(hour)}$}
  \State ($\mathit{array_{cls},Energy_{cls}})=\mathbf{CLS}({\mathit{bestarray}})$
 
  \State $\mathit{bestarray}= 
  \begin{cases}
  \mathit{array_{cls}},& \text{\small if $\mathit{Energy_{cls}>bestEnergy}$}  \\
  \mathit{bestarray},& \text{Otherwise}\\
    \end{cases}
  $\;
   \State Update $\mathit{bestEnergy}$
  \EndWhile
\State \textbf{return} $\mathit{bestarray},\mathit{bestEnergy}$  \Comment{Final Layout and Energy}
\EndProcedure
\end{algorithmic}
\end{algorithm}

\begin{figure}[htbp]
    \centering
    \includegraphics[width=\linewidth]{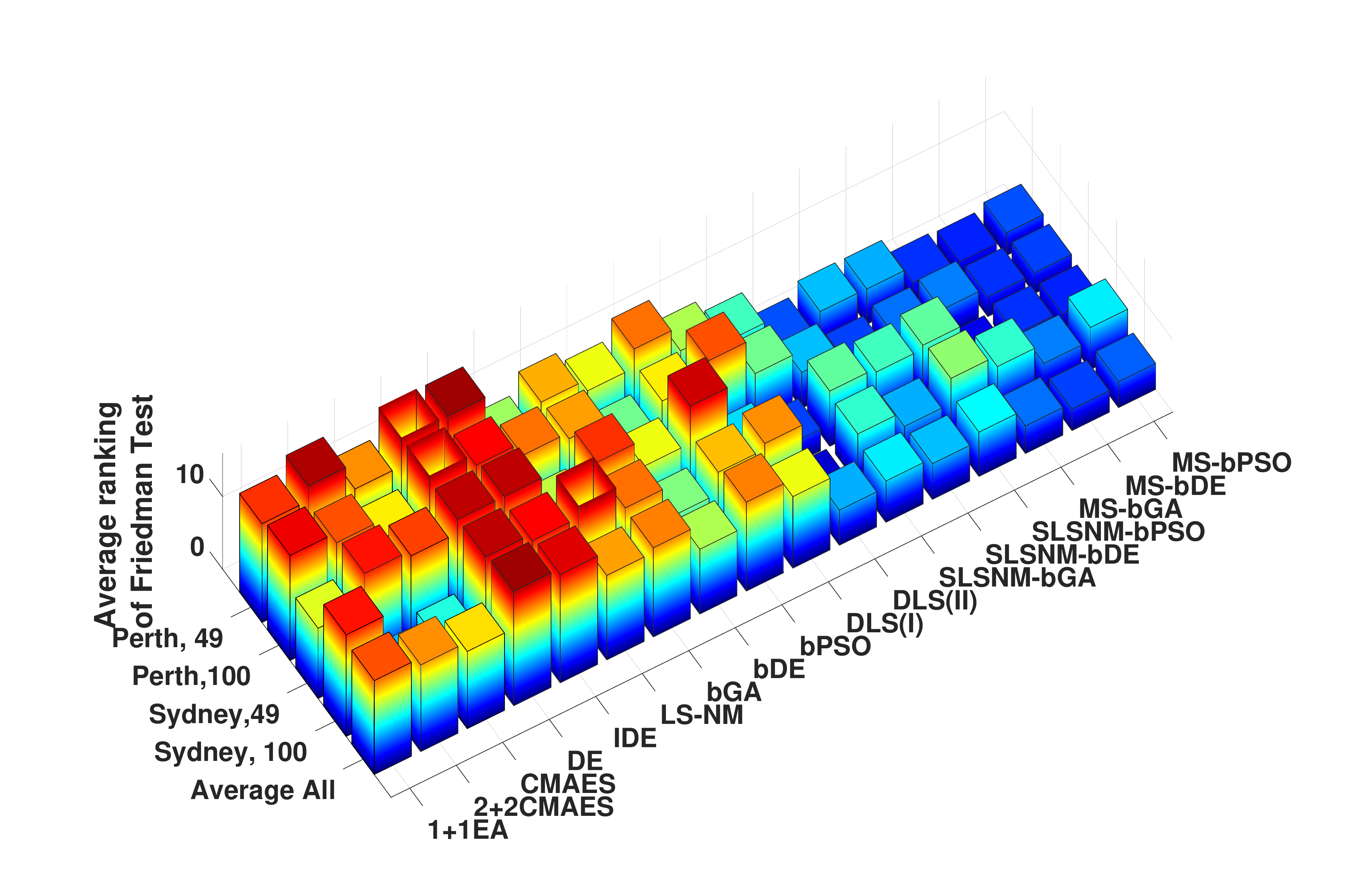}
    \caption{Average ranking of the Friedman test for performance of the proposed optimisation methods. Among all applied heuristic methods, MS-bDE achieves the best average rank in both wave scenarios and wave farm sizes (2.96).  }
    \label{fig:Friedman_Test}
\end{figure}
\begin{figure}[htbp]
    \centering
    \includegraphics[width=0.5\linewidth]{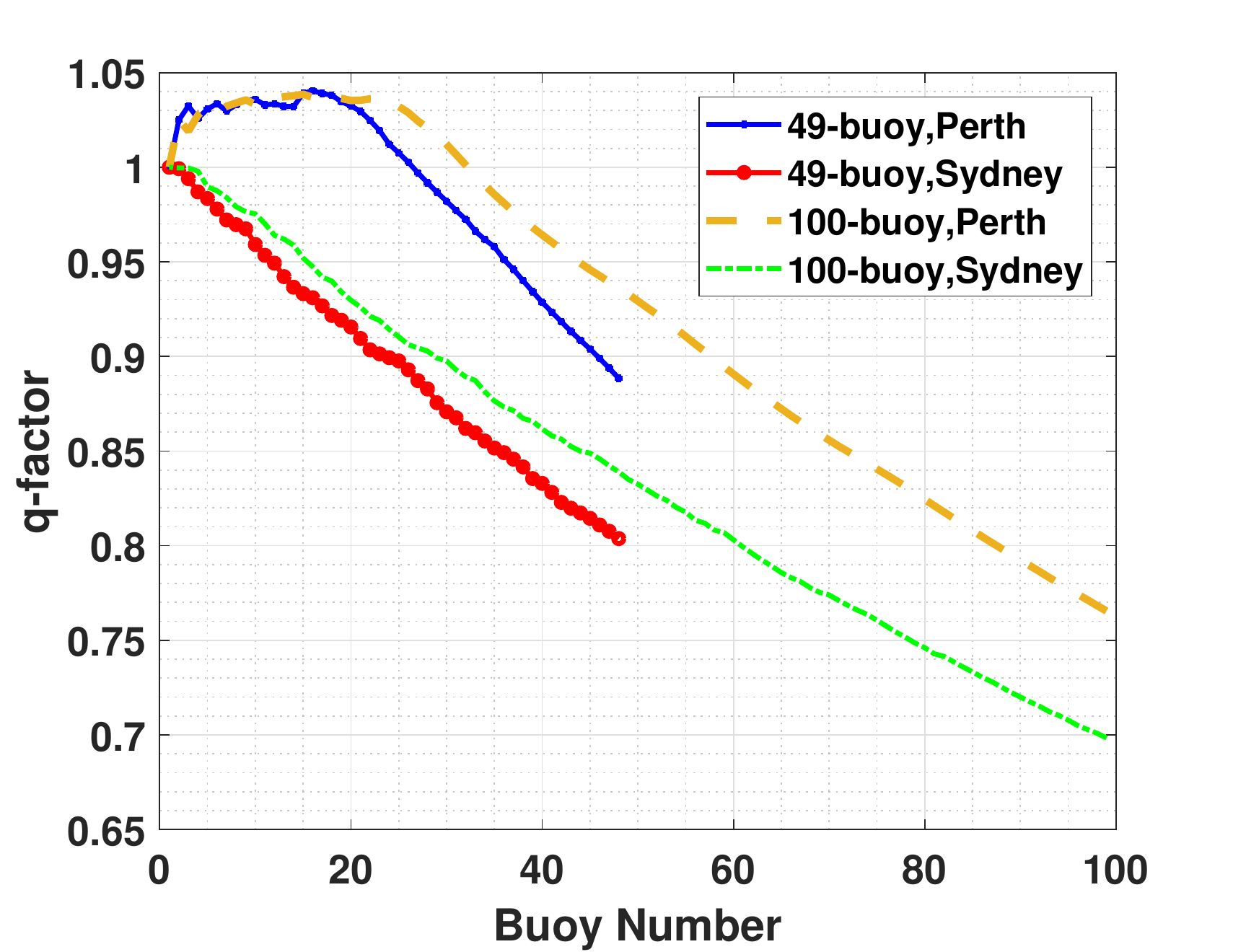}
    \caption{Evaluation of the q-factor performance of the best 49 and 100-buoy layouts by iteratively removing the buoy with the lowest produced power.  }
    \label{fig:q-factor}
\end{figure}

\begin{figure*}[htbp]
    \centering
 \includegraphics[height=7cm]{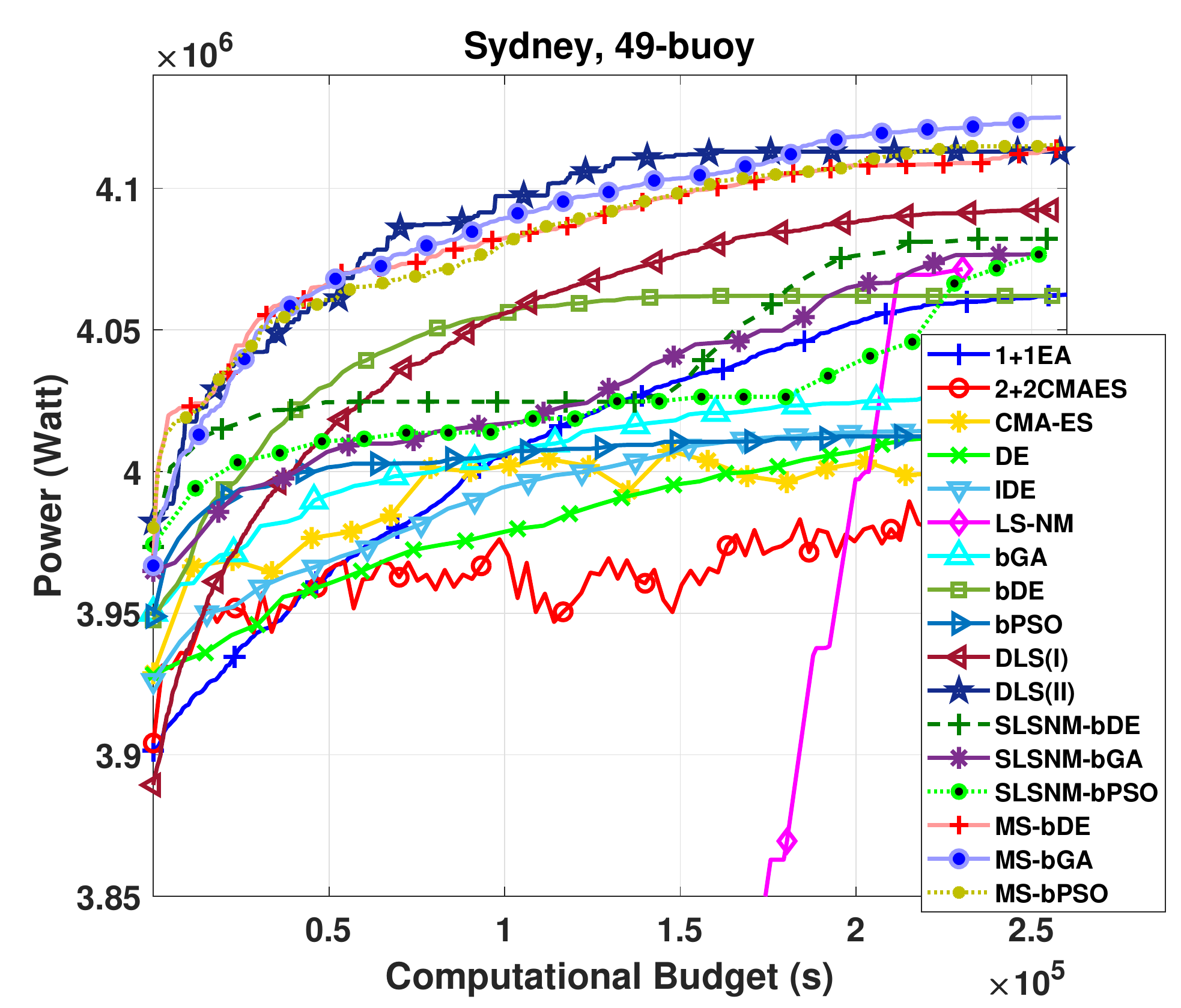}
  \includegraphics[height=7cm]{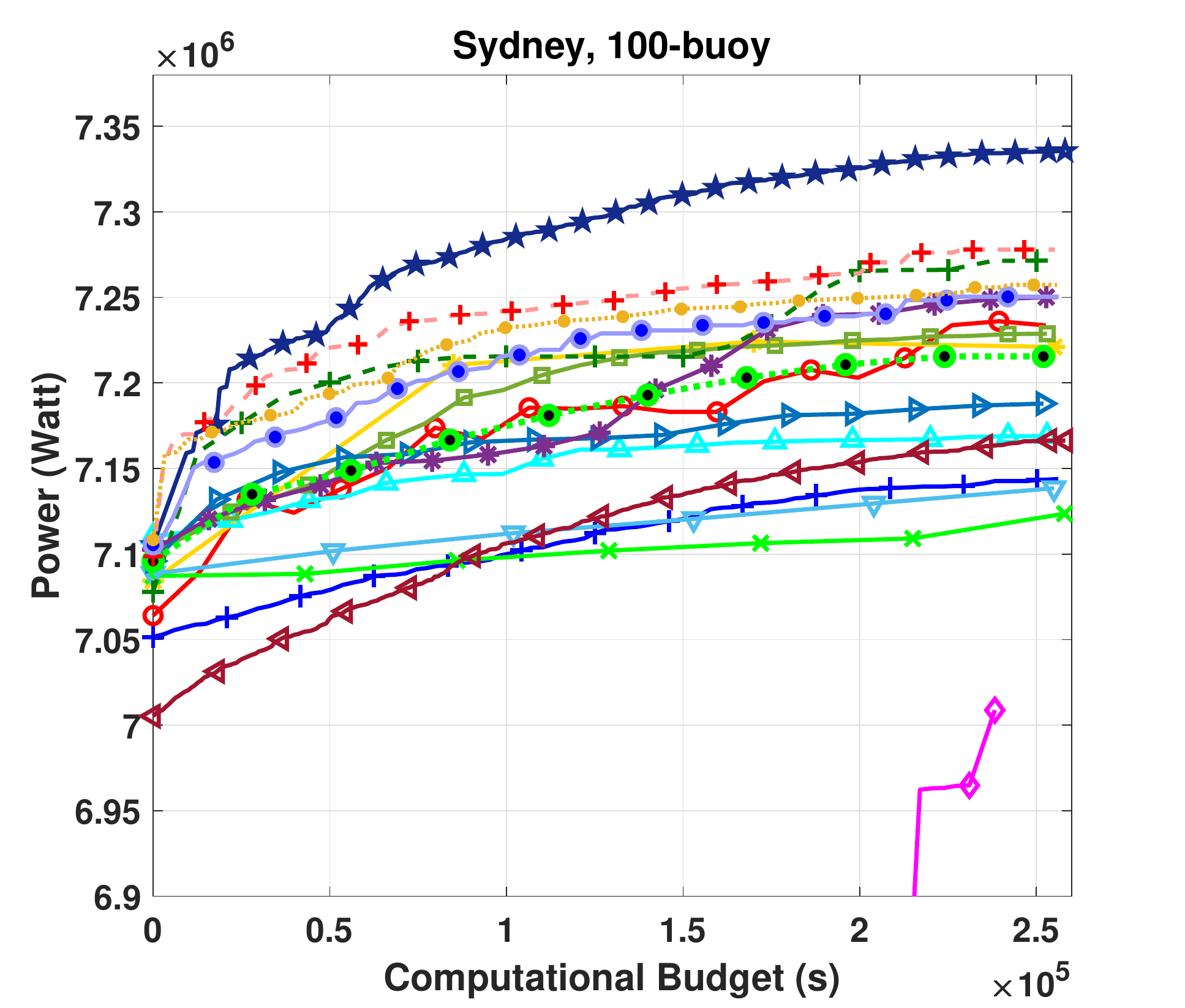}\\
    \includegraphics[height=7cm]{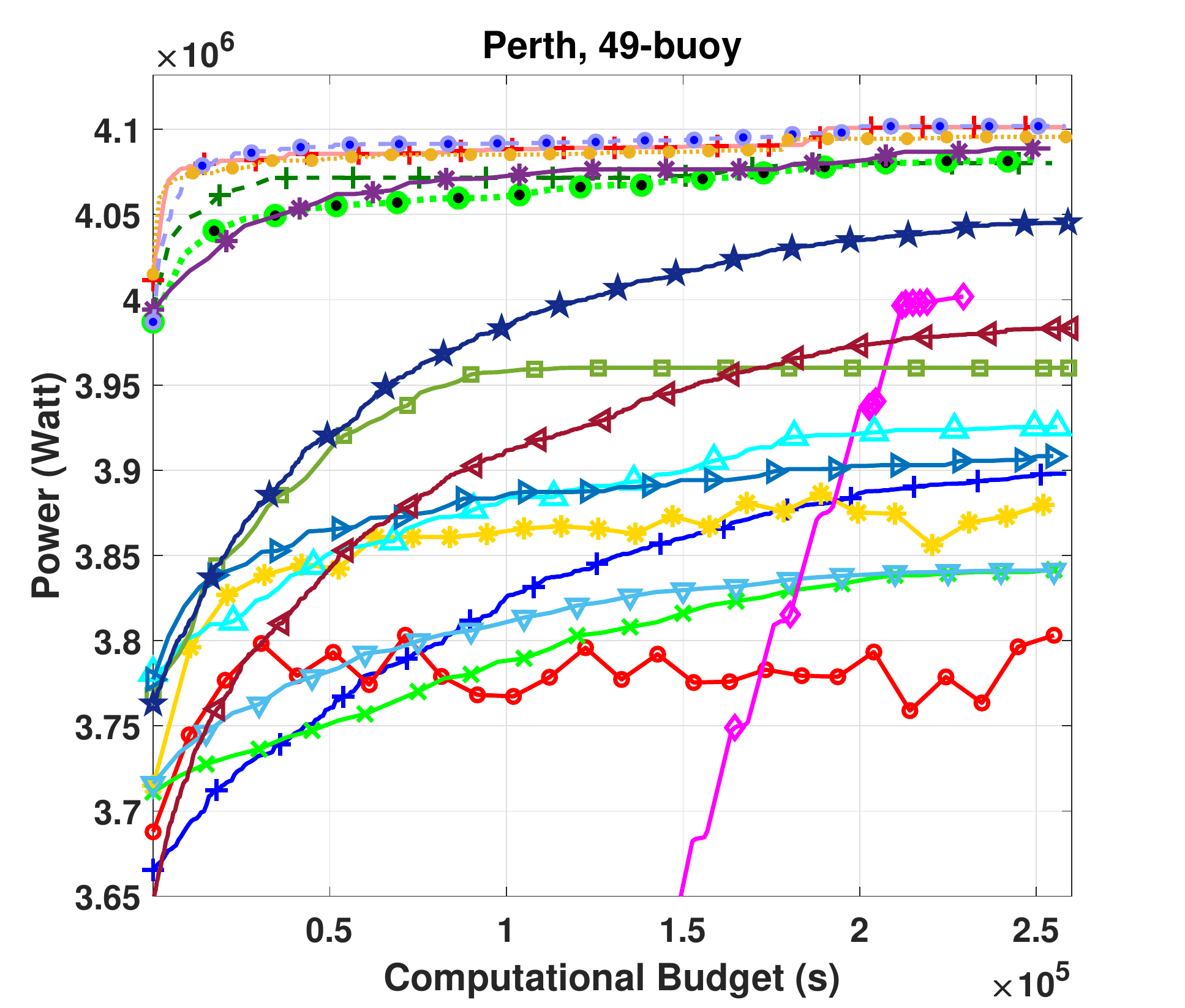}
    \includegraphics[height=6.8cm]{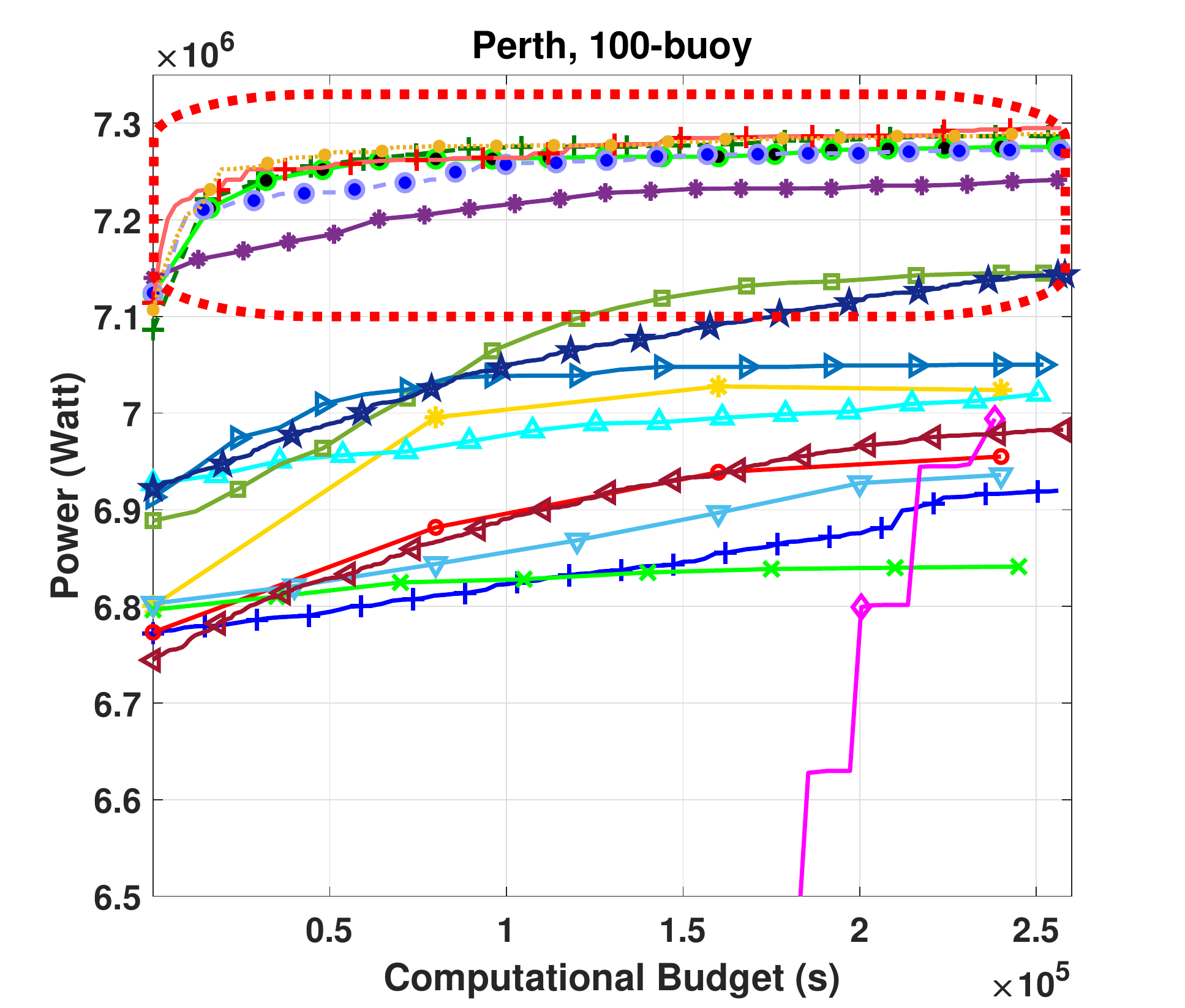}\llap{\raisebox{0.6cm}{\includegraphics[height=3.4cm]{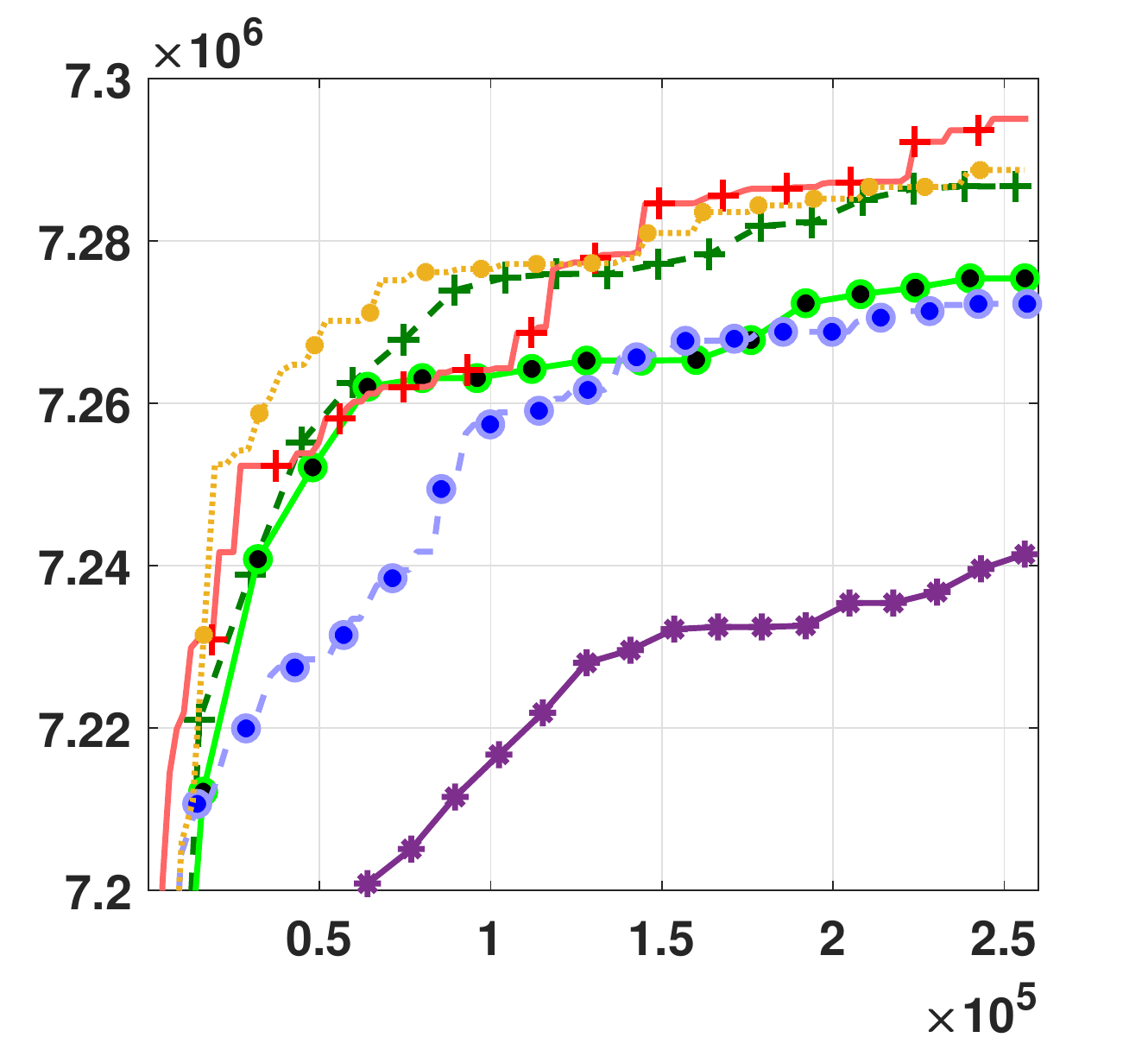}}}
     \caption{Evolution and convergence rate of the average power output of the 17 algorithms for two real wave models. A zoomed version of the plots is provided to give a better view of convergence speed of the new proposed algorithms.}
    \label{fig:Convergence_rate}
\end{figure*}
\begin{sidewaystable}
\setlength{\tabcolsep}{2pt}
 \def\arraystretch{1.05}%
\caption{A performance comparison of the tested heuristics for the 49 and 100-buoy cases, based on maximum, minimum, median and mean power output layout of the best solution per experiment.}
\label{table:allresults}
\scalebox{0.79}{
\begin{tabular}{l|l|l|l|l|l|l|l|l|l|l|l|l|l|l|l|l|l}
\hlineB{4}
\hlineB{4}
\multicolumn{18}{ c }{\textbf{\begin{large}Perth wave scenario (49-buoy)\end{large}}}     \\
\hlineB{2}
 & \textbf{1+1EA}    & \textbf{2+2CMAES}  & \textbf{CMAES}    & \textbf{DE}       & \textbf{IDE}      & \textbf{LS-NM}      & \textbf{bGA}      & \textbf{bDE}      & \textbf{bPSO}     & \textbf{DLS(I)}     & \textbf{DLS(II) }    & \textbf{SLSNM-bGA} & \textbf{SLSNM-bDE} & \textbf{SLSNM-bPSO} & \textbf{MS-bGA} & \textbf{MS-bDE} & \textbf{MS-bPSO} \\
\hline

\texttt{\textbf{Max}}  & 3933532  & 3863159   & 3946189  & 3860444  & 3875783  & 4080796   & 3984477  & 4010788  & 3951612  & 4029731  & 4075129  & 4112549   & 4097756   & 4112602    & 4143453      & \textbf{4177659}     & 4153355       \\
\texttt{\textbf{Min}}  & 3822270  & 3819043   & 3900198  & 3817360  & 3774422  & 3870593   & 3884613  & 3914666  & 3889598  & 3916594  & 4034768  & 4071398   & 4055033   & 4065949    & 4080411     & 4108205      & 4081240      \\
\texttt{\textbf{Mean}} & 3897954  & 3847996   & 3920466  & 3841577  & 3841125  & 4001900   & 3925487  & 3960060  & 3912453  & 3984165  & 4045675  & 4092000   & 4080173   & 4082473    & 4101887      & 4101260      & 4095598       \\
\texttt{\textbf{Median}} & 3904627  & 3853485   & 3923422  & 3842720  & 3848962  & 4008882   & 3919409  & 3964299  & 3912960  & 3987689  & 4044025  & 4090929   & 4081414   & 4082183    & 4099826      & 4095057      & 4089500       \\
\texttt{\textbf{Std}}  & 32160 & 14659  & 17904 & 13118 & 26134 & 57673  & 30027 & 27647 & 17177 & 33047 & 10639 & 10535  & 14291  & 12556   & 15447   & 33245   & 19794    \\ 
\hlineB{4}
\multicolumn{18}{ c }{\textbf{\begin{large}Perth wave scenario (100-buoy)\end{large}}}       \\
\hlineB{2}
 & \textbf{1+1EA}    & \textbf{2+2CMAES}  & \textbf{CMAES}    & \textbf{DE}       & \textbf{IDE}      & \textbf{LS-NM}      & \textbf{bGA}      & \textbf{bDE}      & \textbf{bPSO}     & \textbf{DLS(I)}     & \textbf{DLS(II) }    & \textbf{SLSNM-bGA} & \textbf{SLSNM-bDE} & \textbf{SLSNM-bPSO} & \textbf{MS-bGA} & \textbf{MS-bDE} & \textbf{MS-bPSO} \\
\hline
\texttt{\textbf{Max}}  & 6949622  & 7159987   & 7106268  & 6884148  & 7071418  & 7235426   & 7069638  & 7205581  & 7115011  & 7147428  & 7192402  & 7293928   & 7334975   & 7317723    & 7337150      & \textbf{7347403}      & 7323919       \\
\texttt{\textbf{Min}}  & 6869548  & 6893283   & 6840548  & 6822275  & 6830504  & 6522058   & 6974571  & 7031121  & 7001017  & 6881014  & 7096030  & 7198081   & 7185982   & 7201029    & 7209405     & 7240886      & 7233752       \\
\texttt{\textbf{Mean}} & 6909165  & 6976394   & 7038973  & 6841196  & 6936048  & 6926550   & 7019850  & 7145057  & 7050040  & 6982706  & 7144278  & 7252075   & 7286727   & 7275411    & 7272215     & 7293645      & 7287221       \\
\texttt{\textbf{Median}}& 6908770  & 6931793   & 7058938  & 6834645  & 6912788  & 6888912   & 7017830  & 7160194  & 7036276  & 6978218  & 7143539  & 7250198   & 7294572   & 7276700    & 7259815      & 7287454      & 7286829       \\
\texttt{\textbf{Std}}  & 23522 & 101719 & 68757 & 19143 & 80329 & 213896 & 28773 & 46510 & 33182 & 74593 & 29736 & 26291  & 40569  & 30945   & 48338   & 30656   & 23286    \\
\hlineB{4}
\multicolumn{18}{ c }{\textbf{\begin{large}Sydney wave scenario (49-buoy)\end{large}}}       \\
\hlineB{2}
 & \textbf{1+1EA}    & \textbf{2+2CMAES}  & \textbf{CMAES}    & \textbf{DE}       & \textbf{IDE}      & \textbf{LS-NM}      & \textbf{bGA}      & \textbf{bDE}      & \textbf{bPSO}     & \textbf{DLS(I)}     & \textbf{DLS(II) }    & \textbf{SLSNM-bGA} & \textbf{SLSNM-bDE} & \textbf{SLSNM-bPSO} & \textbf{MS-bGA} & \textbf{MS-bDE} & \textbf{MS-bPSO} \\
\hline
\texttt{\textbf{Max}} & 4082524  & 4036152   & 4061528  & 4028337  & 4062984  & 4089731   & 4052984  & 4078389  & 4036728  & 4108751  & 4134622  & 4105401   & 4100089   & 4084955    & \textbf{4145252}      & 4132385     & 4125968      \\

\texttt{\textbf{Min}} & 4046311  & 4013471   & 4015919  & 4002550  & 3976301  & 4039525   & 4006283  & 4050349  & 3992467  & 4070321  & 4098276  & 4031291   & 4047364   & 4066979    & 4106192      & 4097087      & 4107365       \\

\texttt{\textbf{Mean}} & 4062828  & 4026230   & 4029509  & 4014852  & 4014611  & 4063963   & 4028394  & 4062048  & 4015029  & 4092373  & 4116149  & 4075606   & 4082176   & 4076665    & 4125113      & 4113961     & 4117115      \\

\texttt{\textbf{Median}} & 4060505  & 4029230   & 4030342  & 4015482  & 4011081  & 4064273   & 4025310  & 4059954  & 4015661  & 4095077  & 4113040  & 4072725   & 4083311   & 4076621    & 4127949      & 4113672      & 4117134      \\

\texttt{\textbf{Std}} & 9580  & 7808   & 12500 & 8591  & 20268 & 18106  & 13725 & 7924  & 11643 & 10140 & 10814 & 19897  & 13306  & 4946   & 12113   & 9324   & 5856   \\

\hlineB{4}
\multicolumn{18}{ c }{\textbf{\begin{large}Sydney wave scenario (100-buoy)\end{large}}}          \\
\hlineB{2}
 & \textbf{1+1EA}    & \textbf{2+2CMAES}  & \textbf{CMAES}    & \textbf{DE}       & \textbf{IDE}      & \textbf{LS-NM}      & \textbf{bGA}      & \textbf{bDE}      & \textbf{bPSO}     & \textbf{DLS(I)}     & \textbf{DLS(II) }    & \textbf{SLSNM-bGA} & \textbf{SLSNM-bDE} & \textbf{SLSNM-bPSO} & \textbf{MS-bGA} & \textbf{MS-bDE} & \textbf{MS-bPSO} \\
 \hline

\texttt{\textbf{Max}}  & 7143849  & 7325364   & 7292118  & 7179529  & 7195009  & 7182442   & 7209740  & 7285926  & 7229868  & 7246878  & \textbf{7362279}  & 7288882   & 7338191   & 7290958    & 7300092      & 7337922      & 7309508       \\
\texttt{\textbf{Min}} & 7117313  & 7225156   & 7088371  & 7094498  & 7114283  & 6770261   & 7128255  & 7132653  & 7159246  & 7048150  & 7307167  & 7210217   & 7240062   & 7167618    & 7209436      & 7247446      & 7203600      \\
\texttt{\textbf{Mean}}  & 7143849  & 7268166   & 7242833  & 7123588  & 7138670  & 7008764   & 7168951  & 7228751  & 7187912  & 7166332  & 7335497  & 7246161   & 7271553   & 7228570    & 7250348      & 7277918      & 7257887      \\
\texttt{\textbf{Median}}& 7140541  & 7263653   & 7266642  & 7121216  & 7132516  & 7025996   & 7164768  & 7236813  & 7183483  & 7172586  & 7339777  & 7245065   & 7267610   & 7237676    & 7233001      & 7274685      & 7255276       \\
\texttt{\textbf{Std}} & 18069 & 34055  & 64716 & 24211 & 25094 & 131454 & 25026 & 41309 & 20879 & 57189 & 18211 & 22690  & 31592  & 42732   & 32945   & 26909   & 33889 \\
\hlineB{4}
\hlineB{4}
\end{tabular}
}
\end{sidewaystable}
\begin{figure*}[htbp]
    \centering
    \includegraphics[width=\linewidth]{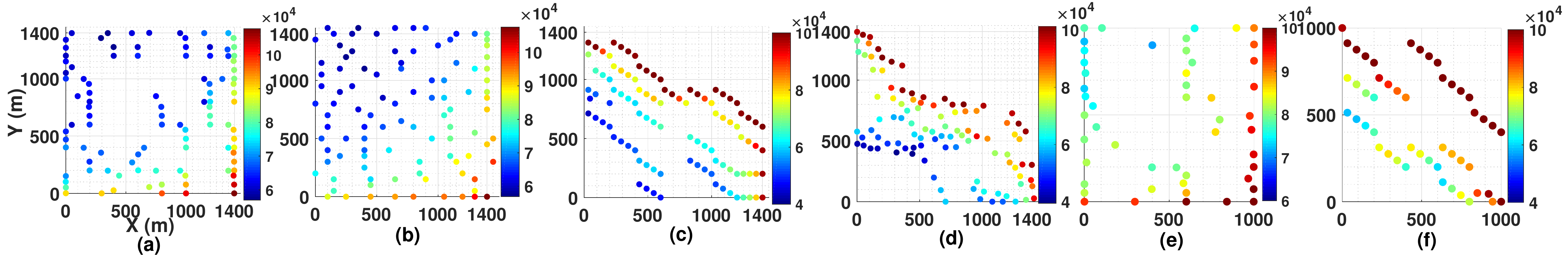}
    \caption{The best 49 and 100-buoy layouts: 
    (a) Power=7337922 (Watt), q-factor=0.7 by MS-bDE for 100-buoy in Sydney wave scenario; 
    (b) Power=7362279 (Watt), q-factor=0.70 for 100-buoy in Sydney wave farm ; 
    (c) Power=7347403 (Watt), q-factor=0.76 by MS-bDE for 100-buoy in Perth wave farm; 
    (d) Power=7235425 (Watt), q-factor=0.75 by LS-NM for 100-buoy in Perth; 
    (e) Power=4145252 (Watt), q-factor=0.80 by MS-bGA for 49-buoy in Sydney ; 
     (f) Power=4177658 (Watt), q-factor=0.88 by MS-bDE for 49-buoy in Perth wave farm (2.4\% more power than LS-NM best 49-buoy layout).
    }
    \label{fig:best_layout}
\end{figure*}
In the Perth wave scenario, the best 49 and 100-buoy layouts are found by MS-bDE. However, we can see the MS-bGA and DLS  perform better than other optimisation methods in the Sydney wave regime for 49 and 100-buoy layouts, respectively.

In addition, Figure~\ref{fig:Friedman_Test} depicts a broad comparison of all proposed optimisation methods by the average ranking of the non-parametric Friedman's test~\cite{hollander2013nonparametric} including both real wave scenarios with two different farm sizes and the total average rank of each method in all case studies. It can be seen that, overall, MS-bDE produces the best optimisation performance. 

In Figure \ref{fig:Convergence_rate}, in all configurations of the Perth wave model, three hybrid and three multi-strategy methods converge very fast and still outperform the other methods. It is notable that these six proposed methods start the optimisation process with a high power output solution due to the smart initialisation technique described in Section \ref{sec:Hybrid}. Looking more closely at Figure~\ref{fig:Convergence_rate}, we can see that all discrete optimisation approaches converge faster than the continuous algorithms on average. Furthermore, because of the embedding of the rotation operator with the binary EAs, the multi-strategy techniques are able to converge faster than the hybrid methods, especially in the initial iterations. In terms of other algorithms, in the Sydney wave model, the performance of the DLS is strong ($N=100$) and outperforms other methods in terms of the convergence rate and the produced power. However, we can see that in the smaller farm, the performance of multi-strategy EAs are competitive, and  MS-bGA performs better than other optimisation methods in the final iterations. 

Some of the most productive 49 and 100-buoy layouts are presented by Figure~\ref{fig:best_layout} from all the runs in the two scenarios. The absorbed power of buoys is characterised by their colour. It can be seen that the best layouts in Perth are multi-row diagonal arrangements; however, this trend is different in the Sydney wave site where the optimisation method pushes some buoys to the farm boundaries.

Lastly, to further investigate the hydrodynamic interactions between buoys in the best layouts, we perform two different analyses. 

In the first analysis, we iteratively remove the buoy with the lowest absorbed power and evaluate the performance of the layout. While this experiment focuses on the least-performing buoy, the interactions of these buoys might be beneficial for the wave farms nevertheless. Figure~\ref{fig:q-factor} shows that a lot of constructive interference is exploited in both the 49 and 100 buoy Perth scenario (up to the 26th buoy), while the marginal improvement from adding buoy's declines after that. For Sydney, there is an almost uniform decline in marginal performance from the start.

\begin{figure*}[htbp]
    \centering
 \includegraphics[height=6cm]{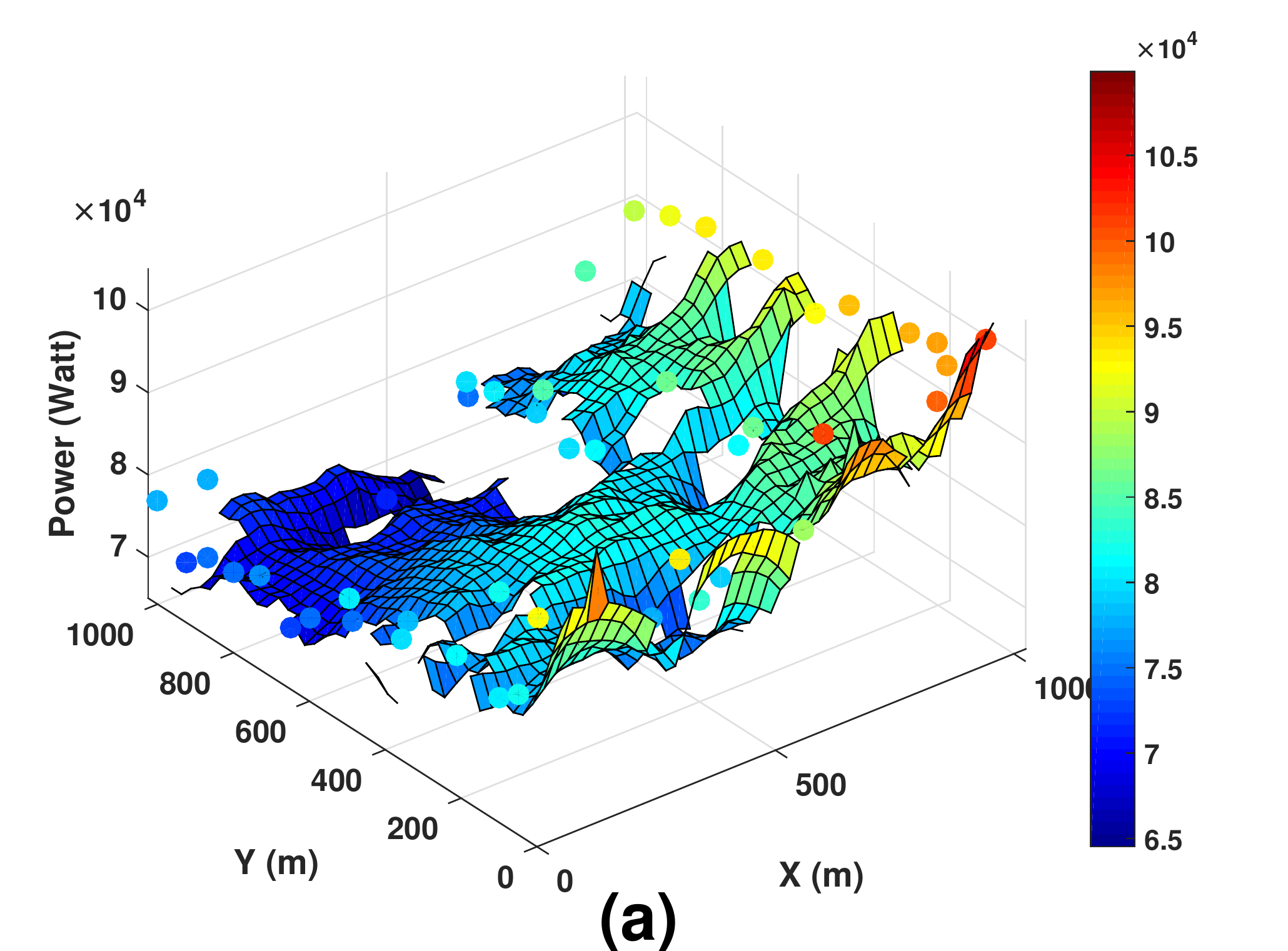}
  \includegraphics[height=6cm]{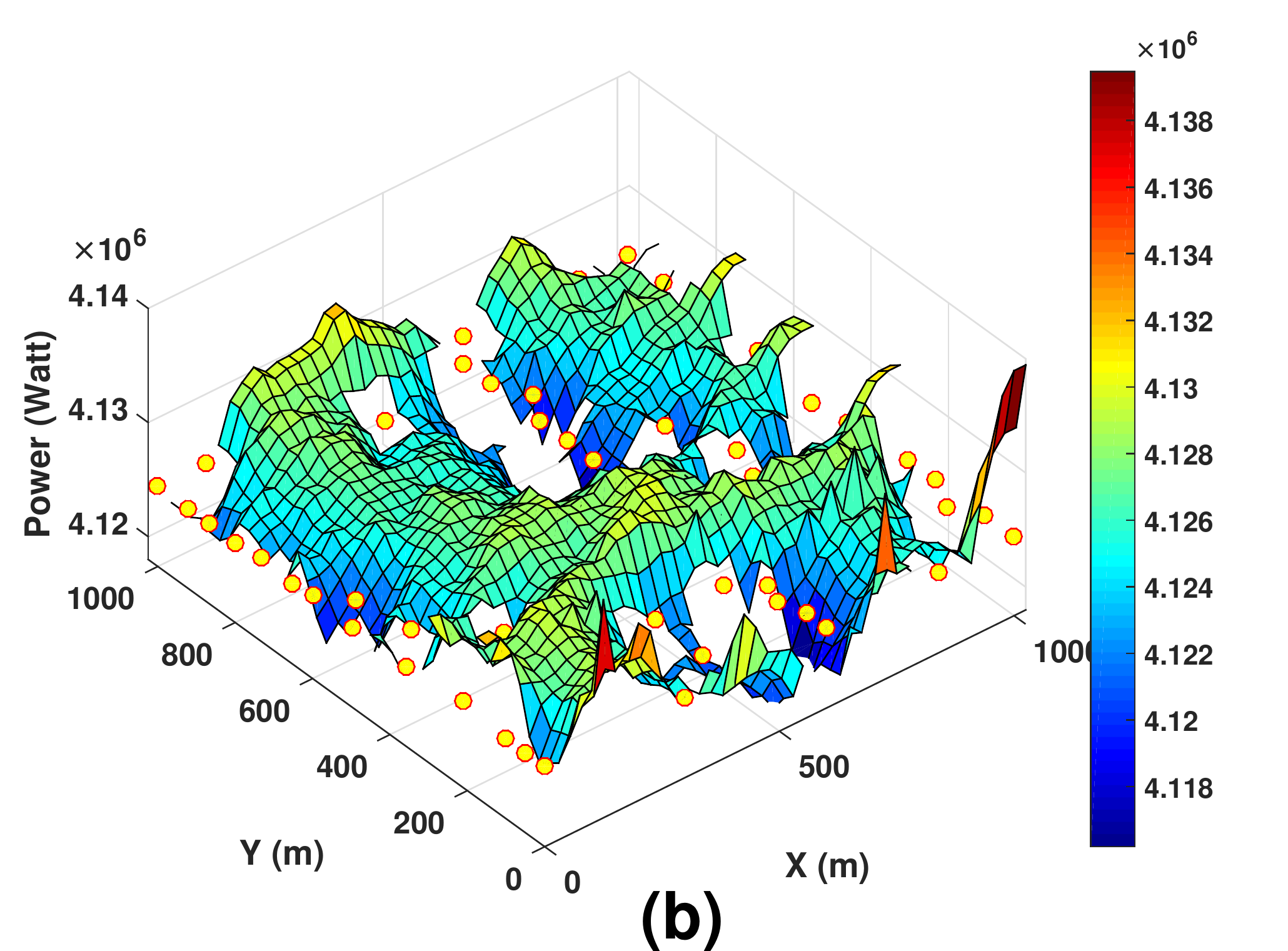}\\
     \includegraphics[height=6cm]{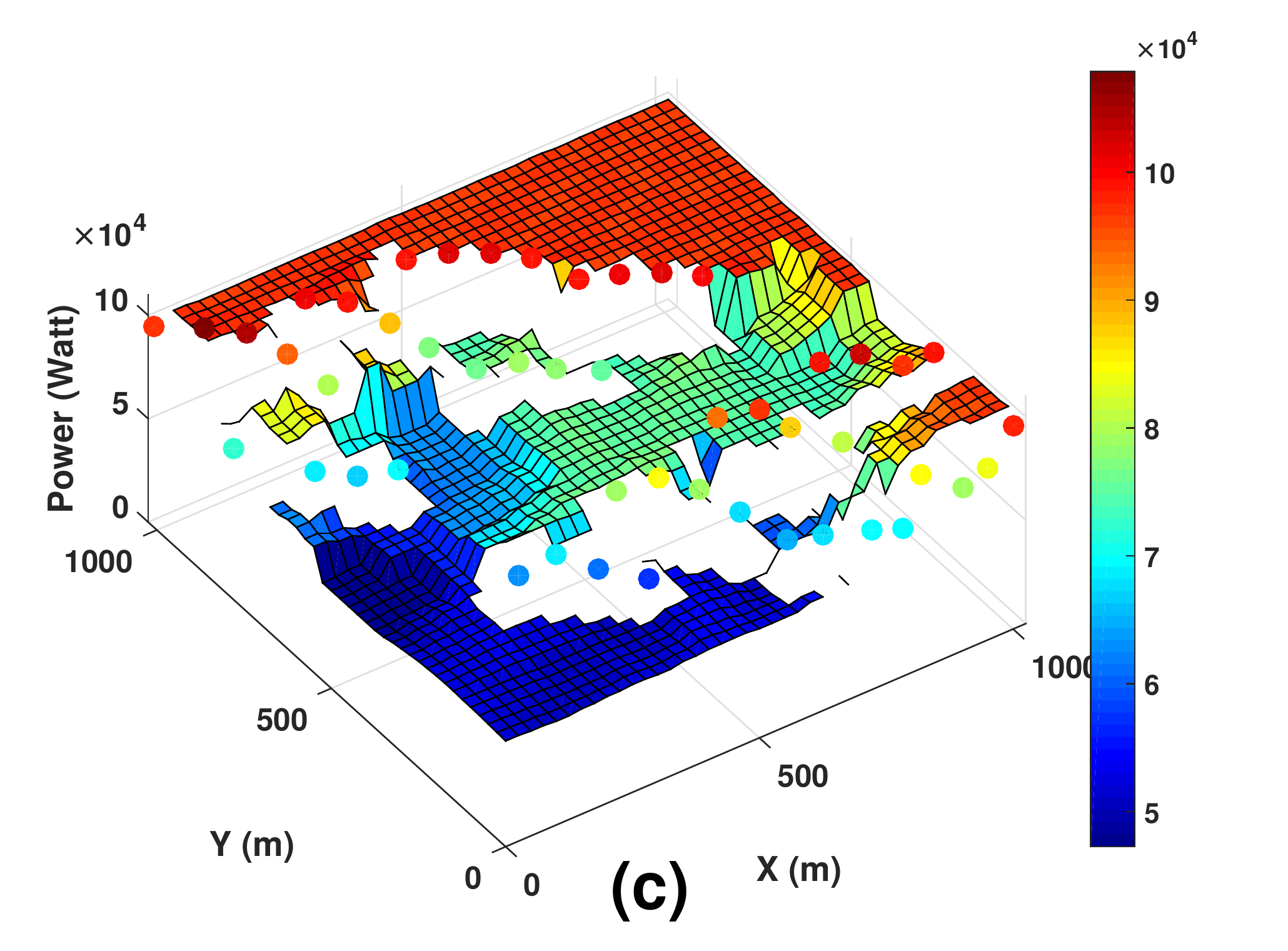}
     \includegraphics[height=6cm]{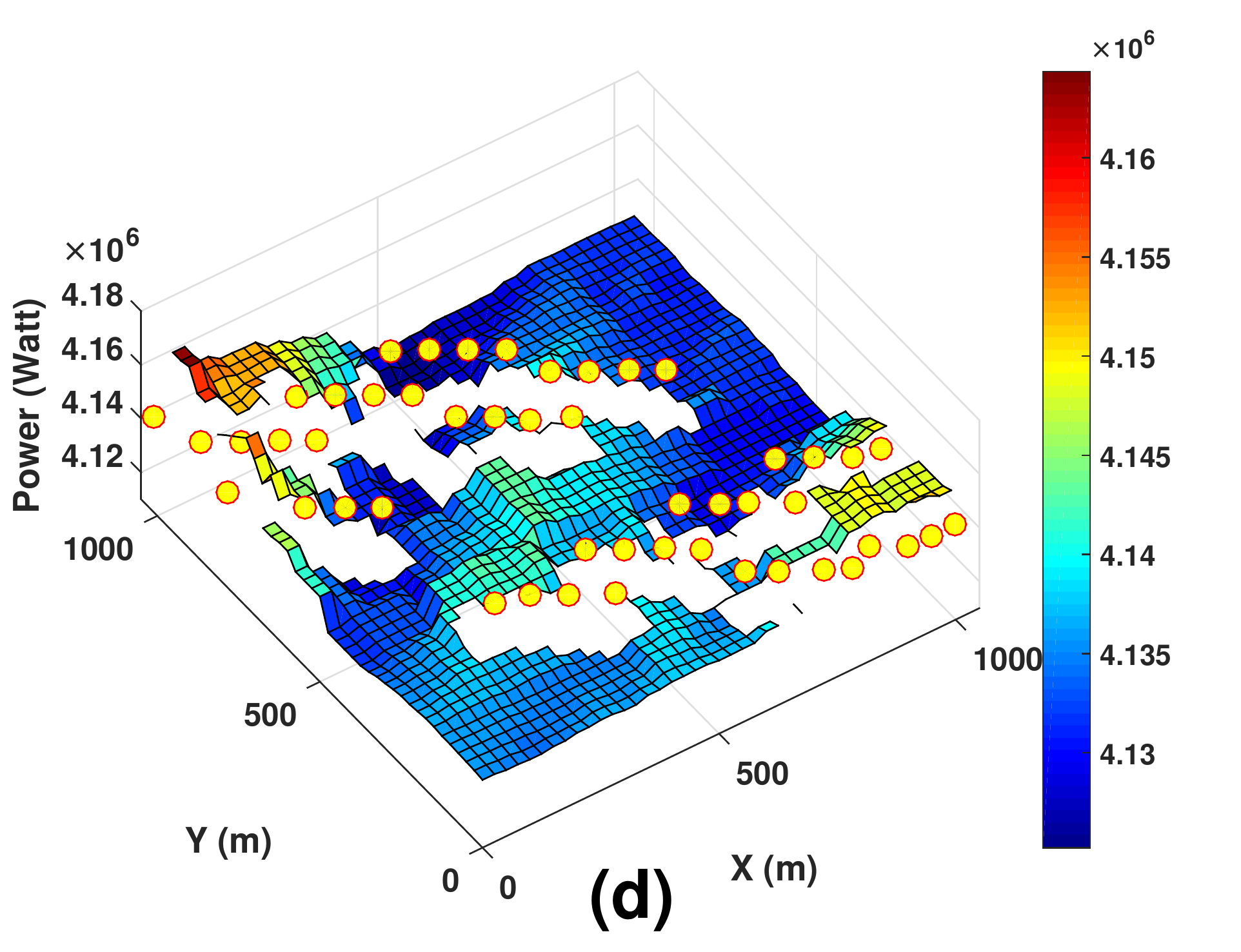}
   
     \caption{Power landscape analysis of the best 49-buoy layouts. (a) Sydney - wave
state - grid sampling the power extracted by the final buoy. (b) Sydney, {\em{total}} energy extracted with grid sampling of the final buoy position  (c) Perth-wave state - grid sampling of last buoy's power. (d) grid sampling of total power w.r.t last buoy's position. No samples are made within the safe distance of already-placed buoys. The power of each buoy is characterised by a specific colour in both (a) and (c).    
}
    \label{fig:power_landscape}
\end{figure*}

The second analysis of the best layouts selects the buoy with the highest power, removes it, and then maps the landscape using a 25-meter grid. 
We record both the absorbed power of the buoy and the total wave farm power output per each sample. Figure~\ref{fig:power_landscape} shows the power landscape analysis of this experiment. 
 
Note that the gaps are the infeasible areas around the already-placed buoys. 
The subplots (b) and (d) indicate a multimodal and complex power landscape, for the placement of the last of the 49 buoys, especially for Sydney. 


\begin{figure*}[htbp]
    \centering
 \includegraphics[height=5.5cm]{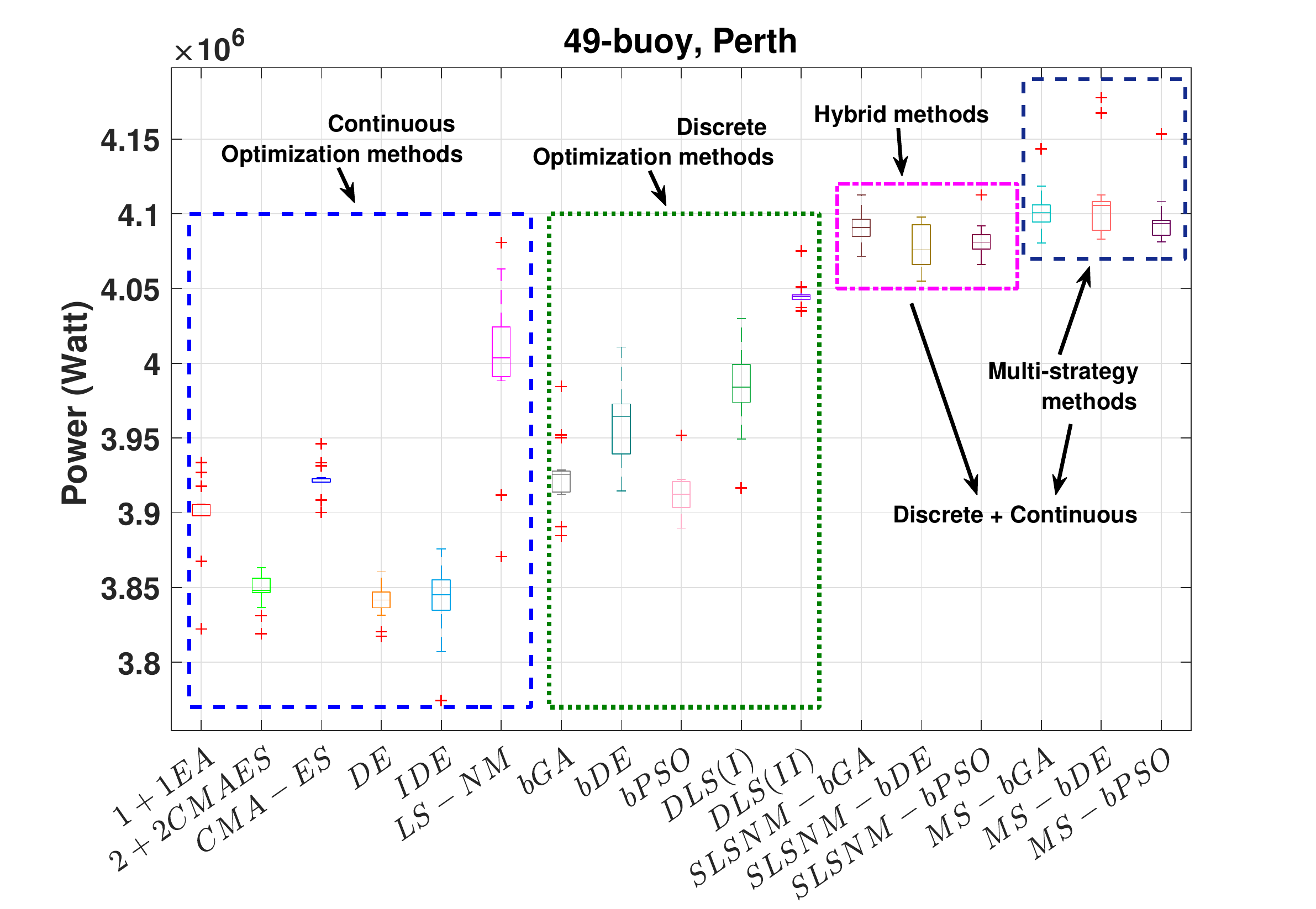}
  \includegraphics[height=5.5cm]{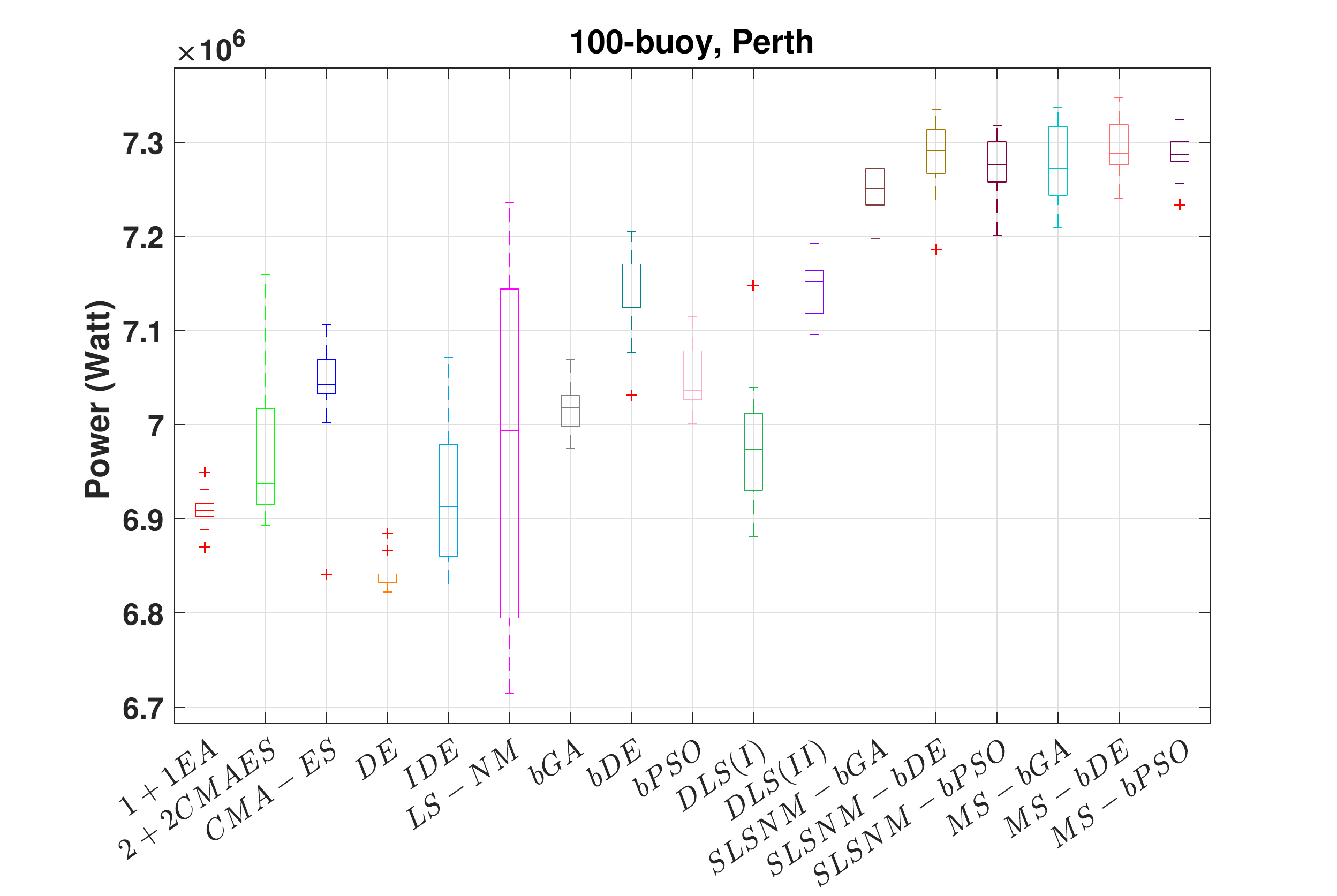}\\
     \includegraphics[height=5.5cm]{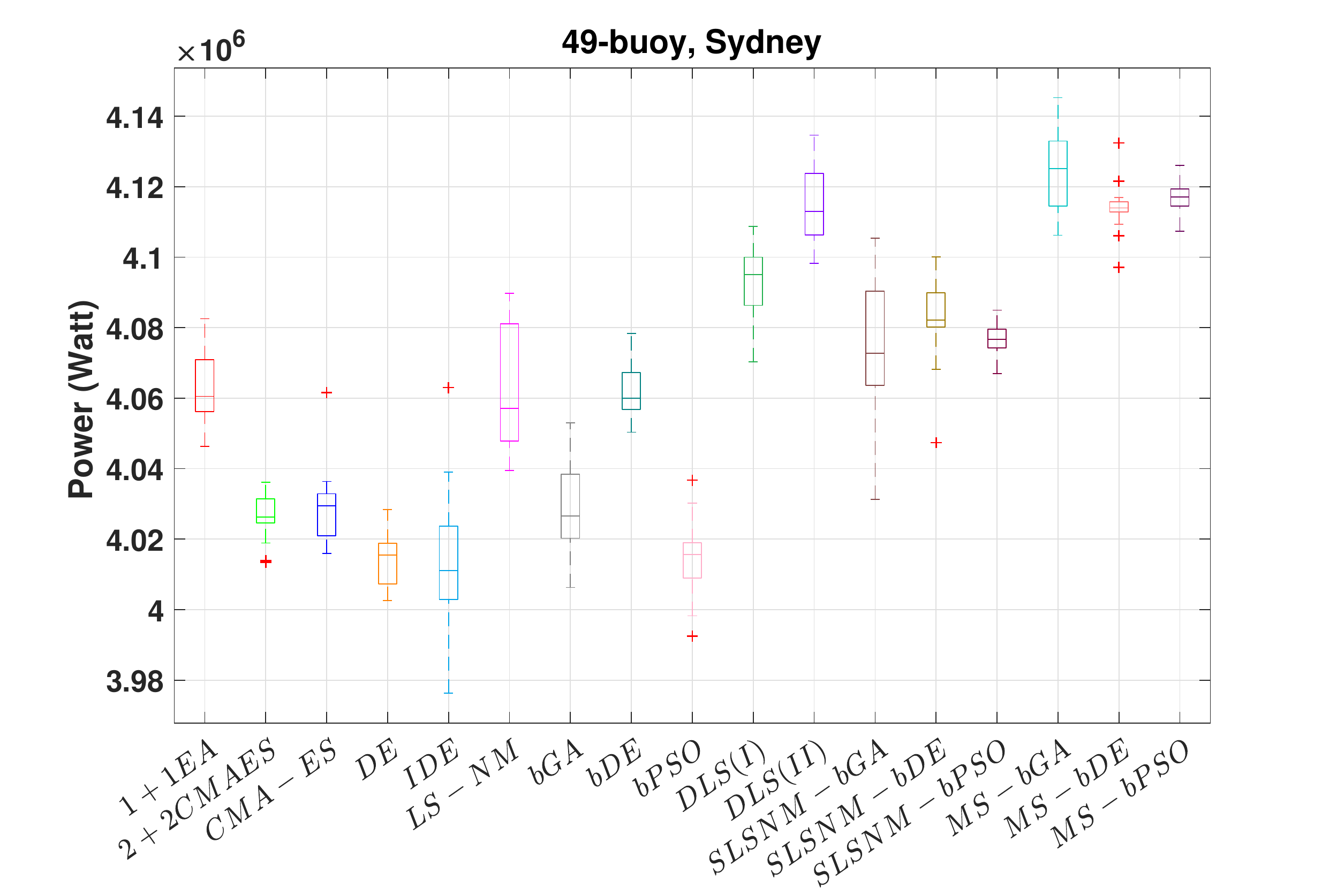}
     \includegraphics[height=5.5cm]{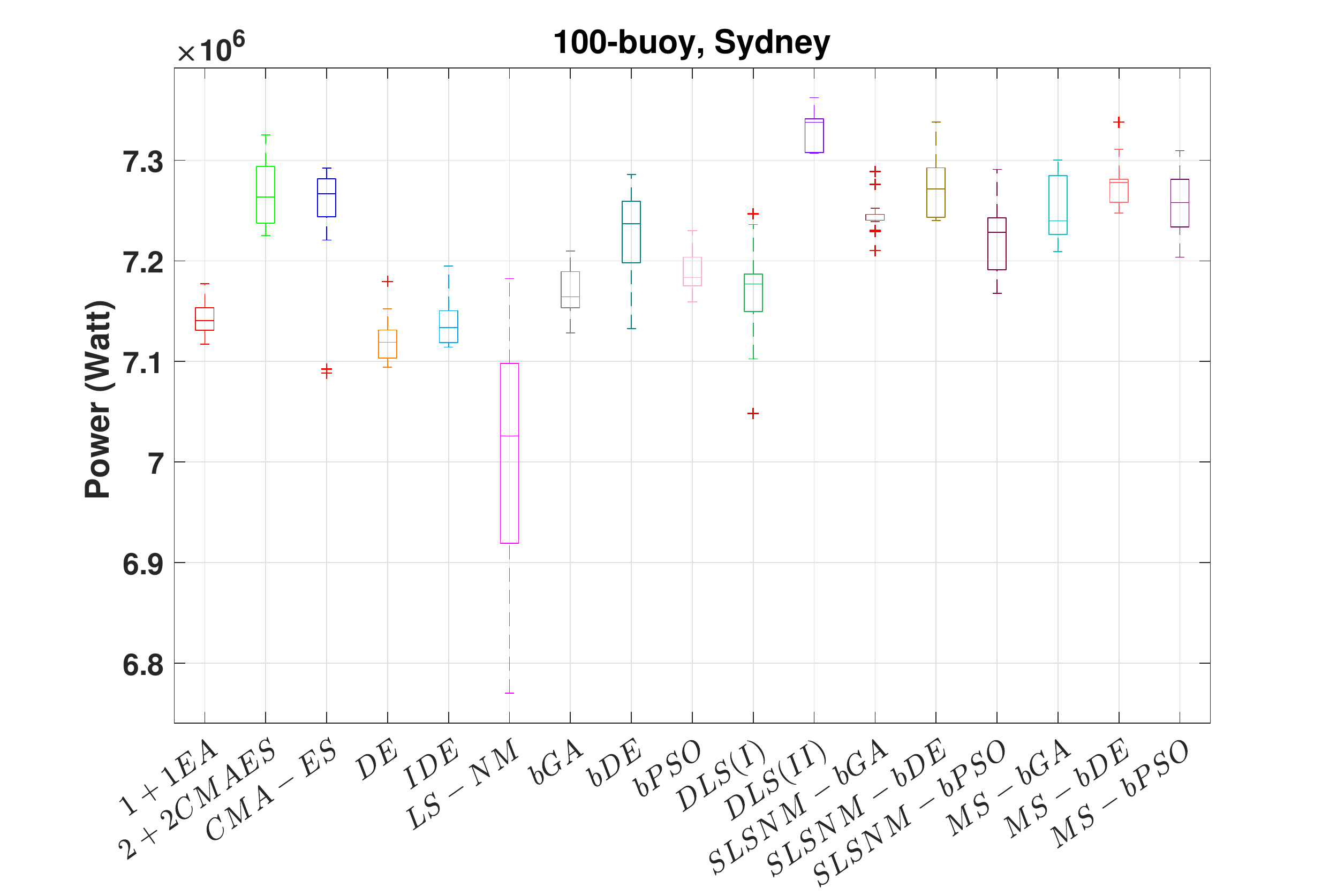}
   
     \caption{ The comparison of the optimisation algorithms performance for 49 and 100-buoy layouts in Sydney and Perth wave models. The optimisation results present the best solution per each experiment. (10 independent runs per each method)}
    \label{fig:boxplot_all}
\end{figure*}

 In order to report on the distribution of the performance of the different approaches across 10 independent runs, the box-plots (Figure~\ref{fig:boxplot_all}) is represented. Figure ~\ref{fig:boxplot_all} shows and highlights the considerable performance of the proposed multi-strategy optimisation framework compared with other optimisation techniques in the large wave farms problem.   
 
 Meanwhile, All implemented codes and auxiliary materials are publicly available: \url{https://cs.adelaide.edu.au/~optlog/research/energy.php}.
 
\section{Conclusions}\label{sec:conclusions}

In this paper, we have proposed, assessed, and systematically compared 17 different optimisation approaches for optimising the arrangement of large wave farms with 49 and 100 generators in two real wave regimes (Sydney and Perth). This study comprised three new hybrid algorithms, each with three variants as a multi-strategy EA framework customised to this field.

This optimisation problem is challenging in terms of the cost of its evaluation model and the large multimodal search landscape. 
Our new framework addresses this problem through careful problem decomposition into sub-farms, the use of discrete search spaces and a customised mutation operator (rotation). 

The statistical results indicate that the new multi-strategy evolutionary algorithm consisting of symmetric local search and Nelder Mead search, combined with an embedded rotation operator, plus an improved binary DE and a hybrid backtracking strategy (DLS+CLS) performs better than other applied optimisation methods on average. In our experiments, this method overcomes other state-of-the-art algorithms, for both 49 and 100-buoy layouts, in terms of convergence speed and power production.
 
Future work could explore other optimisation dimensions, including considering other effective buoy designs and power take-off system settings. 
\section*{Acknowledgements}
We would like to offer our special thanks to Dr.Hansen,  Dr.kalami and Dr.Mirjalili for sharing and publishing their valuable source codes and also thank Mr.Roostapour for his good comments. Additionally, This research is supported with supercomputing resources provided by the Phoenix HPC service at the University of Adelaide.
\bibliographystyle{unsrt}  
\bibliography{sample-bibliography}
\end{document}